\documentclass{article}

\PassOptionsToPackage{numbers,sort&compress}{natbib}
\usepackage[preprint]{neurips_2026}

\usepackage[utf8]{inputenc}
\usepackage[T1]{fontenc}
\usepackage{hyperref}
\usepackage{url}
\usepackage{booktabs}
\usepackage{amsmath,amsfonts,amssymb}
\usepackage{bm}
\usepackage{algorithmic}
\usepackage{algorithm}
\usepackage{float}
\usepackage{array}
\usepackage{textcomp}
\usepackage{graphicx}
\usepackage{xcolor}
\usepackage{cancel}
\usepackage[caption=false,font=footnotesize]{subfig}
\usepackage[normalem]{ulem}
\usepackage{empheq}
\usepackage{multirow}
\usepackage{amsthm}

\newtheorem{theorem}{Theorem}
\newtheorem{definition}{Definition}

\newtheorem{corollary}{Corollary}
\newcommand{\authorblock}[3]{%
  \begin{tabular}[t]{c}
    \textbf{#1}\\
    \normalfont #2\\
    {\normalfont\footnotesize\texttt{#3}}
  \end{tabular}%
}
\usepackage{wrapfig}
\allowdisplaybreaks[4]
\title{FedVSSAM: Mitigating Flatness Incompatibility in Sharpness-Aware Federated Learning}
\author{%
\normalfont
\begin{tabular}{@{}c@{\hspace{1.5em}}c@{\hspace{1.5em}}c@{}}
\authorblock{Bingnan Xiao}{Fudan University}{22110720061@m.fudan.edu.cn} &
\authorblock{Yuan Gao}{Fudan University}{23110720072@m.fudan.edu.cn} &
\authorblock{Bingcong Li}{ETH Zurich}{bingtsongli@gmail.com}
\end{tabular}\\[1.25em]
\begin{tabular}{@{}c@{\hspace{4em}}c@{}}
\authorblock{Wei Ni}{Edith Cowan University}{wei.ni@ieee.org} &
\authorblock{Xin Wang}{Fudan University}{xwang11@fudan.edu.cn}
\end{tabular}\\[1.25em]
\authorblock{Tony Q. S. Quek}{Singapore University of Technology and Design}{tonyquek@sutd.edu.sg}
}

\begin{document}
\maketitle
\begin{abstract}

Sharpness-aware minimization (SAM) is an effective method for improving the generalization of federated learning (FL) by steering local training toward flat minima. Under data heterogeneity, however, device-side SAM searches for locally flat basins that are incompatible with the flat region preferred by the global objective. We identify this structural failure mode as \emph{flatness incompatibility}, which explains why improving local flatness alone may provide limited training and generalization improvement for the global model. We reveal that flatness incompatibility arises from data heterogeneity and the friendly adversary phenomenon, and is further amplified by local updates and partial device participation. To mitigate this issue, we propose Federated Learning with variance-suppressed sharpness-aware minimization (FedVSSAM), which constructs a variance-suppressed adjusted direction and uses it consistently in local flatness search, local descent, and global update. FedVSSAM anchors both perturbation and update directions to a more stable global direction, instead of correcting only an isolated local perturbation. We establish non-convex convergence guarantees of FedVSSAM and prove that the mean-square deviation between the adjusted direction and the global gradient is effectively controlled. Experiments demonstrate that FedVSSAM mitigates flatness incompatibility and outperforms the baselines across diverse FL settings.

\end{abstract}

\section{Introduction}
Federated learning (FL) enables devices to train collaboratively while keeping personal data local to protect privacy.~\cite{9060868,9347706}. 
In a classical FL paradigm such as FedAvg~\cite{fedavg}, selected devices perform multiple local updates and then send their models to the parameter server (PS) for aggregation. 
This communication-efficient training paradigm, however, is highly sensitive to data heterogeneity, i.e., non-independent and identically distributed (non-IID) data across devices. 
Under non-IID data, local objectives lead different devices toward different regions of the parameter space. 
The resulting device drift leads to divergence between local and global updates, thereby slowing down convergence~\cite{li2019convergence}.
Moreover, increasing the number of local training iterations, which are used to reduce communication load, exacerbates the divergence~\cite{9835537}.

Improving generalization is another central objective of FL, since the global model should perform well across different device distributions. 
Sharpness-aware minimization (SAM) is a natural tool for this purpose. 
Instead of solely minimizing the loss at the current parameter, SAM seeks locally flat solutions by constructing an adversarial perturbation within a neighborhood and updating the model with the gradient evaluated at the perturbed parameters~\cite{foret2021sharpnessaware}.
Following this principle, FedSAM applies SAM to local training to improve the generalization of the global model under FL~\cite{caldarola2022improving}.

However, the inherent challenge of SAM-based FL is not merely applying SAM to local objectives. Due to data heterogeneity, the flat region searched by a device can be substantially incompatible with the flat region required by the aggregated global model. 
This local--global inconsistency is severe because the SAM perturbation direction determines both the neighborhood used for local flatness search and the point where the subsequent descent gradient is evaluated. 
Once the local perturbation is misaligned with the global loss landscape, the error is propagated from perturbation construction to perturbed local descent and is accumulated through aggregation. 
Therefore, improving local flatness alone may still drive the aggregated model away from a globally flat basin, especially under strong data heterogeneity and multiple local training iterations~\cite{Shi_2023_CVPR,pmlr-v235-lee24aa}.

We term this structural failure mode \emph{flatness incompatibility}, which is induced by two sources: (i) data heterogeneity, which makes local gradients and local flat regions deviate from the global loss landscape; and (ii) the friendly adversary phenomenon, where a perturbation that is adversarial to one batch may become benign to another~\cite{li2024enhancing}. 
These two sources do not enter SAM-based FL as isolated errors. Data heterogeneity biases the local gradient and local flatness search, while friendly adversary makes the perturbation direction sensitive to the particular batch data.
The resulting mismatch affects both the perturbation direction and the point where the descent gradient is evaluated. 
As training progresses, such mismatch can accumulate and enter the aggregated update. 
Thus, correcting only one training stage reduces part of the error, but could not keep the local flatness search, local update, and global update aligned. This motivates using a shared and stable direction throughout training to stabilize flatness search and model updates under flatness incompatibility.

This paper proposes Federated Learning with variance-suppressed sharpness-aware minimization (FedVSSAM) to improve global generalization and model training of SAM-based FL under data heterogeneity. 
The key idea is to construct a variance-suppressed adjusted direction and use it in local flatness search, local update, and global update consistently, to anchor the perturbation and update directions to a more stable global direction.
Our key contributions are summarized as follows: 
\begin{itemize}
    \item 
    We first identify flatness incompatibility in SAM-based FL and explicitly reveal its two sources, i.e., data heterogeneity and friendly adversary. This theoretical characterization turns the generalization issue of SAM-based FL into a concrete consistency problem among local flatness search, local update, and global update.

    \item 
    To address flatness incompatibility, we develop FedVSSAM to provide a shared and stable direction across local flatness search, local update, and global update, thereby mitigating flatness incompatibility and promoting consistent loss landscapes across devices.

    \item 
    We derive non-convex convergence guarantees for FedVSSAM, accounting for flatness incompatibility and partial device participation, and quantify how the designed variance suppression operations tighten the MSE between the adjusted direction and the global gradient.
    Our analysis reveals that FedVSSAM achieves a superior convergence by only transmitting updated models to the PS with no additional auxiliary vectors for the uplink transmission, e.g., control or dual variables, compared with existing SAM-based FL algorithms.

    \item 
    Extensive experiments validate that FedVSSAM can effectively combat flatness incompatibility and accelerate system convergence under data heterogeneity and partial device participation. 
    It is also shown that the designed variance suppression contributes positive performance gains and that FedVSSAM remains robust to key system hyperparameters.
\end{itemize}

\subsection{Related Work}
Inherent data heterogeneity in FL leads to device drifts and impedes training~\cite{kairouz2021advances,zhao2018federated,hsu2019measuring}. FedProx, SCAFFOLD, and FedDyn mitigate device drifts via proximal penalties, control variates, or dynamic regularization~\cite{fedprox,scaffold,feddyn}. FedCM and FedACG further stabilize local and global updates through momentum-based corrections~\cite{xu2021fedcm,kim2024communication}. Such methods mainly reduce optimization bias induced by data heterogeneity, but do not explicitly target FL generalization from a flatness perspective.

To enhance generalization, SAM seeks flat minima by minimizing a worst-case neighborhood loss~\cite{foret2021sharpnessaware}, which is motivated by previous generalization studies~\cite{hochreiter1997flat,keskar2017on,jiang2019fantastic}. Later variants analyze SAM's performance or improved perturbation strategy~\cite{pmlr-v162-andriushchenko22a,pmlr-v139-kwon21b,du2022efficient}. Friendly adversary shows batch-based perturbations can lose adversariality~\cite{li2024enhancing}. However, these works focus on centralized optimization and cannot address perturbation inconsistency in heterogeneous SAM-based FL.

In SAM-based FL, FedSAM first employs SAM to improve FL generalization~\cite{caldarola2022improving}. MoFedSAM extends FedSAM with momentum-based stabilization, while FedGAMMA introduces global sharpness-aware correction with SCAFFOLD~\cite{qu2022generalized,10269141}. FedSMOO and \mbox{FedLESAM} pursue consistent flatness via dynamic regularization or global perturbations~\cite{pmlr-v202-sun23h,pmlr-v235-fan24c}. However, the aforementioned flatness incompatibility exists throughout the SAM-based FL training stage and needs to be processed coherently. Please refer to Appendix~\ref{app:additional-related-work} for a detailed related work summary.

\section{System Model and Preliminaries}

\subsection{SAM-based Federated Learning}
Consider a classical FL framework comprising a PS, and $N$ devices collected by $\mathcal{N} \buildrel \Delta \over = \{ 1, \cdots ,N \}$. The devices train collaboratively a model $\theta \in \mathbb{R}^d$ to minimize the following finite sum problems:
\begin{align} \label{FL obj}
\mathop {\min }\limits_{\theta \in \mathbb{R}^d}  F(\theta ) = \frac{1}{N}\sum\limits_{i \in {\cal N}} {{F_i}} (\theta ) ,\
F_i(\theta) \buildrel \Delta \over =  \mathbb{E}_{\phi_i \sim \mathcal{D}_i} \left[ F_i\left(\theta ; \phi_i\right) \right] ,
\end{align}
where $F: \mathbb{R}^d \rightarrow \mathbb{R}$ is the global objective function at the BS, $F_i$ is the empirical loss of the $i$-th device, and $\phi_i$ denotes a random batch of data sampled from the local dataset $\mathcal{D}_i$ of the $i$-th device, with distributions differing across devices due to data heterogeneity. 

Instead of minimizing $F(\theta)$ only at a single model parameter vector, SAM~\cite{foret2021sharpnessaware} seeks a model parameter vector whose loss remains small under small perturbations of the model in a centralized setting. This favors flat basins of $F(\cdot)$, where the objective varies mildly within a neighborhood of the model, and the resulting model typically has better generalization.
As such, the following minimax problem is formulated:
\begin{align} \label{SAM obj}
    \min _{\theta} \max _{\| {\epsilon} \| \leq \rho}  F(\theta+ {\epsilon}) ,
\end{align}
where $\rho$ denotes the perturbation radius, and $\epsilon$ represents SAM's perturbation.
By adopting the SAM optimizer at each device in a distributed manner, 
{FedSAM aims to improve local flatness and the generalization of the global model in SAM-based FL.}
The objective of FedSAM is formulated as
\begin{align} \label{FedSAM obj}
\min _{\theta} \max _{\left\|\epsilon_i\right\| \leq \rho}\left\{f(\tilde{\theta}) = \frac{1}{N} \sum_{i \in \mathcal{N}} f_i(\tilde{\theta})\right\} ,
\end{align}
where $f(\tilde{\theta}) \! \buildrel \Delta \over =  \! \max _{\|\epsilon \| \leq \rho} F(\theta \!+\! \epsilon)$, and $ f_i(\tilde{\theta}) \! \buildrel \Delta \over = \! \max _{\|\epsilon_i\| \leq \rho} F_i(\theta \!+\! \epsilon_i) $ with $\tilde{\theta} = \theta + \epsilon_i $; $\epsilon_i$ and $\epsilon$ denote the local and global perturbations, respectively.

Constrained by data privacy and infrequent communication between the PS and devices, SAM-based FL algorithms construct the local perturbation $\epsilon_i$ on each device $i$ for generalization.
To efficiently approximate the inner maximization of $f_i(\tilde{\theta})$, we utilize the first-order Taylor expansion, as in prior SAM-based FL studies~\cite{caldarola2022improving,pmlr-v235-lee24aa,qu2022generalized,10269141,pmlr-v202-sun23h,pmlr-v235-fan24c}. The resulting linearized inner maximization of \eqref{FedSAM obj} is
\begin{align} \label{stochastic linearization}
    \epsilon_i 
    \!=\! \mathop {\arg \max }\limits_{\left\| \epsilon_i \right\| \le \rho } F_i(\theta \!+\! \epsilon_i) 
    {\approx} \mathop {\arg \max }\limits_{\| \epsilon_i \| \le \rho } \{ \! F_i ( \theta ) \!+\! \epsilon_i^{\top} \nabla F_i(\theta)  \!+\! \mathcal{O}(\rho^2) \! \} 
    {\approx} \rho \frac{\nabla F_i (\theta)}{ \| \nabla F_i (\theta) \| } 
    {\approx} \rho \frac{ g_i }{ \| g_i \| } , 
\end{align}
where $g_i \!=\! \nabla F_i(\theta ; \phi_i)$ denotes the stochastic gradient computed on batch data $\phi_i$
to alleviate computational burden. See Appendix~\ref{app:baselines} for the detailed training process of FedAvg~\cite{fedavg} and FedSAM~\cite{caldarola2022improving}.

\subsection{Flatness Incompatibility in FedSAM}\label{subsec:flat-incompat}
\begin{figure}
   \centering
   \subfloat[Local Model]{\includegraphics[width=3cm]{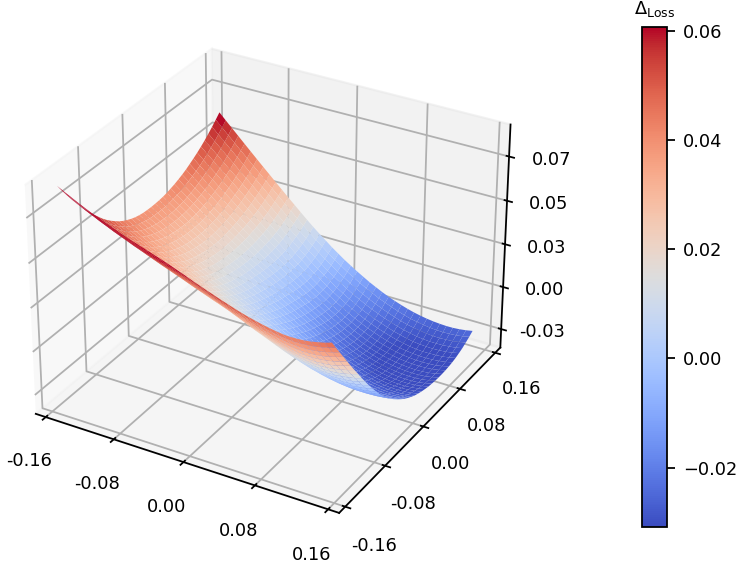}}
   \subfloat[Global Model]{\includegraphics[width=3cm]{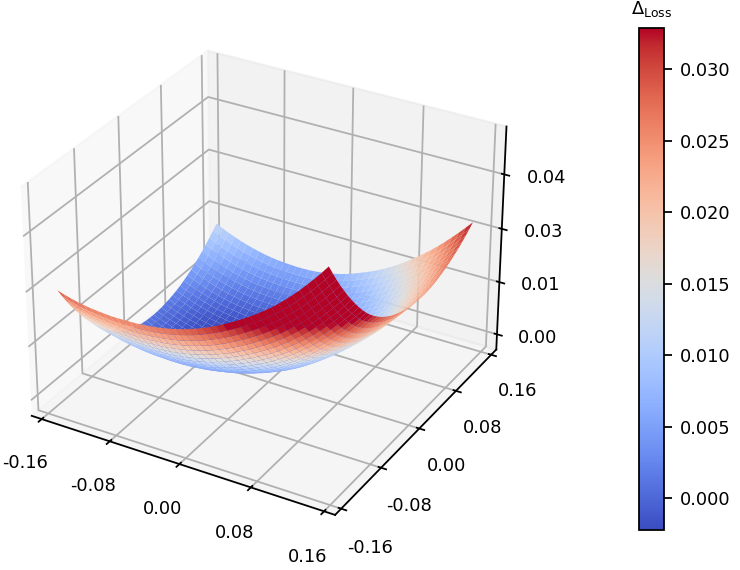}}
  \caption{Visualization of local and global loss landscapes for FedSAM on CIFAR-10. Setup includes 100 devices, 10\% UAR sampling, Dirichlet-0.1, and a 3-layer CNN.}
  \label{fig:fi}
\end{figure}
Fig.~\ref{fig:fi} illustrates flatness incompatibility by comparing the loss landscapes of a local model and the aggregated global model under FedSAM. The low-loss region of the local model is shifted from that of the global model, suggesting that the perturbation $\epsilon_i$ may guide a device toward a locally flat basin that is not aligned with the globally flat basin characterized by $\epsilon$. Hence, improving local flatness alone does not necessarily ensure that the aggregated model remains in a flat region of the global loss landscape. We next define flatness incompatibility, and identify data heterogeneity and friendly adversary as its two sources, which motivates the new variance suppression designs in Section~\ref{vs section}.
\begin{definition}[Flatness incompatibility]
For a perturbation radius $\rho>0$, define the local and global $\rho$-flatness proxies as $s_i(\theta,\rho) = F_i(\theta + \rho \frac{g_i}{\|g_i\|}) - F_i(\theta)$
and $\bar{s}(\theta, \rho) = \frac{1}{N} \sum_{j \in \mathcal{N}} s_j(\theta, \rho)$, respectively.
Flatness incompatibility is quantified by the inter-device dispersion of local flatness proxies:
\begin{align} \label{flatness incompatibility definition}
\Delta_{\mathrm{FI}}(\theta,\rho) &:= \frac{1}{N}\sum_{i \in \mathcal{N}} \Big(s_i(\theta,\rho) - \bar{s}(\theta, \rho) \Big)^2.
\end{align}
A larger $\Delta_{\mathrm{FI}}(\theta,\rho)$ implies that devices perceive mutually incompatible local flat regions at $\theta$ and deviate from globally flat region.
\end{definition}
For SAM-based FL, flatness incompatibility consists of the following two sources, as analyzed in Section~\ref{convergence sec}. One source is data heterogeneity, which can be quantified at $\theta$, as given by
\begin{align} \label{definition delta_i}
    \delta_i(\theta) := \nabla F_i(\theta) - \nabla F(\theta), \quad \forall i \in \mathcal{N},
\end{align}
where $\delta_i(\theta)$ measures the bias of device gradients relative to the global gradient, and a larger $\|\delta_i(\theta)\|^2$ indicates greater data heterogeneity.
Notably, $\delta_i(\theta)$ serves as a proxy for device drift~\cite{li2019convergence}, as it directly quantifies how far the local descent direction deviates from the global descent direction.

The other source of flatness incompatibility is the friendly adversary phenomenon~\cite{li2024enhancing} in SAM-based FL, which refers to the case where the SAM perturbation constructed from one batch of data to increase a device's batch loss can reduce the loss for another batch of data when their stochastic gradients are misaligned.
Consider two different data batches $\phi_i,\phi_i' \sim \mathcal{D}_i$ sampled $\forall i \in \mathcal{N}$, and define $g_i = \nabla F_i(\theta;\phi_i)$, $g_i' = \nabla F_i(\theta;\phi_i')$, and $\epsilon_{i,\phi_i} = \rho \frac{g_i}{\|g_i\|}$, which is the maximizer of the first-order linearized inner problem for batch $\phi_i$.
Based on the first-order Taylor expansion of SAM in \eqref{stochastic linearization}, the local batch loss change of batch $\phi_i'$ along the perturbation constructed from $\phi_i$ satisfies
\begin{align}
\label{friendly_adversary}
F_i(\theta \!+\! \epsilon_{i,\phi_i};\!\phi_i') \!-\! F_i(\theta; \! \phi_i')
\!\approx\! \rho \frac{\langle g_i', g_i\rangle}{\|g_i\|}
\!\le\! 0,
\, \text{if} \ \langle g_i', g_i\rangle \! \le \! 0 .
\end{align}
From \eqref{friendly_adversary}, although $\epsilon_{i,\phi_i}$ is constructed for the linearized inner maximization of batch $\phi_i$, it can reduce the loss of another batch $\phi_i^{\prime}$ when the two stochastic gradients are misaligned.
This indicates that the perturbation obtained from one data batch may fail to faithfully reflect the local landscape of $F_i$ around $\theta$. 
With stochastic gradient variability and data heterogeneity in SAM-based FL, such local perturbations can vary substantially across local updates, thereby causing local flatness search to probe different local flat regions across successive updates.
To quantify the stochastic inconsistency underlying friendly adversary, we define the local gradient deviation as
\begin{align} \label{definition zeta_i}
    \zeta_{i,\phi_i}(\theta) = g_i - \nabla F_i(\theta), \quad \forall i \in \mathcal{N},
\end{align}
with $\mathbb{E}[\|\zeta_{i,\phi_i}(\theta)\|^2]$ giving the batch gradient variance and reflecting the level of instability in~\eqref{stochastic linearization}.

To analyze the sources that affect $\Delta_{\mathrm{FI}}(\theta, \rho)$, we establish the following two lemmas.

\smallskip
\noindent\textbf{Lemma 1.} \textit{Suppose that each local objective function $F_i$ is $L$-Lipschitz smooth. With $\delta_i(\theta)$ and $\zeta_{i,\phi_i}(\theta)$ defined in \eqref{definition delta_i} and \eqref{definition zeta_i}, flatness incompatibility $\Delta_{\mathrm{FI}}(\theta, \rho)$ is upper bounded:}
\begin{align} \label{eq:fi-master}
\mathbb{E}\big[\Delta_{\mathrm{FI}}(\theta,\rho)\big] \! \leq \!
\frac{3\rho^2}{N} \! \sum_{i \in \mathcal{N}} \! \mathbb{E} \! \left[ \|\delta_i(\theta)\|^2 \!+\! 4\|\zeta_{i,\phi_i}(\theta)\|^2 \right] \! +\! \frac{3}{4}L^2\rho^4.
\end{align}
\smallskip
\noindent\textbf{Lemma 2.} \textit{For any fixed $\theta$, sample a batch data sample $\phi_i$ uniformly at random from $\mathcal{D}_i$, $\forall i \in \mathcal{N}$. 
With $g_{i} = \nabla F_{i}(\theta;\phi_i)$, $\delta_i(\theta)$ and $\zeta_{i,\phi_i}(\theta)$ satisfy
}
\begin{align} \label{lemma 2-2}
& \mathbb{E}\big[\,\langle \delta_i(\theta),\,\zeta_{i,\phi_i}(\theta)\rangle\,\big] = 0 \ \text{and} \ \mathbb{E}\Big[\big\| g_i - \nabla F(\theta)\big\|^2\Big]
 = \! \mathbb{E}\Big[\!\big\|\delta_i(\theta)\big\|^2\Big]
\!+\! \mathbb{E}\Big[\!\big\|\zeta_{i,\phi_i}(\theta)\big\|^2\Big]. 
\end{align}
The proofs of \textbf{Lemmas 1} and~\textbf{2} are provided in Appendix~\ref{app:proof-lemma1} and~\ref{app:proof-lemma2}, respectively.
\textbf{Lemma 1} reveals that inter-device data heterogeneity $\delta_i(\theta)$ and intra-device friendly adversary $\zeta_{i,\phi_i}(\theta)$, deteriorate $\Delta_{\mathrm{FI}}(\theta,\rho)$. 
Moreover, an excessively large $\rho$ can amplify $\Delta_{\mathrm{FI}}(\theta,\rho)$ due to incompatible local loss landscapes.
\textbf{Lemma 2} shows $\delta_i(\theta)$ and $\zeta_{i,\phi_i}(\theta)$ are orthogonal in expectation, justifying the insights on the composition of $\Delta_{\mathrm{FI}}(\theta,\rho)$.
Meanwhile, \textbf{Lemma 2} decomposes the deviation between $g_i$ and $\nabla F(\theta)$ into the two additive and non-overlapping sources of flatness incompatibility: data heterogeneity and friendly adversary.
These results indicate that {mitigating flatness incompatibility should not treat $\delta_i(\theta)$ and $\zeta_{i,\phi_i}(\theta)$ as isolated errors.}
As shown in \textbf{Algorithm}~\ref{fedsam and fedavg}, the SAM perturbation determines the perturbed gradient's evaluation point. Therefore, the errors induced by the two sources can affect both the perturbation direction and the subsequent local update, and further enter the global update via aggregation. 
This motivates incorporating one variance-suppressed direction into local flatness search, local update, and global update of SAM-based FL, to \emph{consistently suppress data heterogeneity and friendly adversary of flatness incompatibility throughout the training stage}.

\section{Federated Variance-Suppressed SAM} \label{vssam sec}
This section presents FedVSSAM to tackle flatness incompatibility caused by data heterogeneity and friendly adversary, and improve the global model performance. We start with assumptions.

\noindent \textbf{Assumption 1. (Smoothness)} Each local objective function $F_i$ is $L$-smooth, i.e., 
\begin{align}
    \left\|\nabla F_i(\theta_1)-\nabla F_i(\theta_2)\right\| \leq L\|\theta_1-\theta_2\|, \quad \forall i \in \mathcal{N}.
\end{align}
\noindent \textbf{Assumption 2. (Bounded variance)} For any device $i$ and $ \theta \in \mathbb{R}^d$, the stochastic gradient $ \nabla F_i(\theta, \phi )$ is unbiased with bounded variance, i.e.,
\begin{align}
    \mathbb{E} \big[ \nabla F_i(\theta;\phi) \big]=\nabla F_i (\theta) \ \text{and}\ \mathbb{E}\left[ \| \nabla F_i(\theta;\phi)-\nabla F_i\left(\theta \right) \|^2 \right] \leq \sigma_{l,1}^2 , \quad \forall i \in \mathcal{N}.
\end{align}
\noindent \textbf{Assumption 3. (Bounded heterogeneity)} The global variability of $\nabla F_i(\theta), \forall i$, is bounded, i.e.,
\begin{align}
\mathbb{E}\left[\left\|\nabla F_i\left(\theta\right)-\nabla F\left(\theta\right)\right\|^2 \right] \leq \sigma_{g,1}^2, \quad \forall i \in \mathcal{N}.
\end{align}

\textbf{Assumptions 1--3} are standard assumptions considered in non-convex optimization and FL~\cite{fedprox,Shi_2023_CVPR,scaffold,xu2021fedcm,feddyn,moon,reddi2021adaptive,fednova,kingma2014adam,pmlr-v235-lee24aa,qu2022generalized,10269141,pmlr-v202-sun23h,pmlr-v235-fan24c}. \textbf{Assumptions 2} and \textbf{3} capture the measurement of the friendly adversary phenomenon and data heterogeneity, respectively, as delineated in Section~\ref{subsec:flat-incompat}.

\begin{algorithm}[!t]
    \caption{FedVSSAM 
    }
    \label{vssam fixed gamma and tau}
    \begin{algorithmic}[1]
    \STATE {\textbf{Input:} $T$, $K$, $N$, $\theta^0$, $h^{0}=0$, $\rho$, $\eta_l$, $\eta_g$, $\gamma_l$, $\gamma_g$ }
    \FOR{$t=0,1,\dots T-1$}
        \STATE {The PS selects the device subset $\mathcal{S}^t \subseteq \mathcal{N}$ UAR, and sends ${\theta}^{t}$ and $h^{t}$ to the selected devices}
        \FOR{each device $i \in \mathcal{S}^t$ in parallel}
            \STATE { Initialize local model $\theta_i^{t,0} = {\theta}^t$ and restore ${\theta}^t$}
            \FOR{$k=0,1,\dots K-1$}
                \STATE { Compute gradient $g_i^{t,k} =  \nabla F_i(\theta_i^{t,k};\phi_i^{t,k})$; Perform local flatness search with \eqref{vs flatness search}}
                \STATE {Compute gradient $\tilde{g}_i^{t,k} = \nabla F_i(\tilde{\theta}_i^{t,k}; \phi_i^{t,k})$; Perform local update with \eqref{vs local update}}
            \ENDFOR
            \STATE {Set $ \theta_i^{t} = \theta_i^{t,K}$ and send $\theta_i^{t}$ to the PS}
        \ENDFOR
        \STATE {The PS obtains $g^{t+1} = \frac{1}{SK}\sum_{i \in \mathcal{S}^t}\sum_{k=0}^{K-1} u_i^{t,k}$, and updates $\theta^t$ to $\theta^{t+1}$ by \eqref{vs global update}}
    \ENDFOR
    \RETURN{${{\boldsymbol{\theta}}^{T}}$}
    \end{algorithmic}
\end{algorithm} 

\subsection{Variance Suppression Design of FedVSSAM} \label{vs section}

While it is feasible to generate the local perturbation $\epsilon_i^{t,k}$ and perform local SAM updates via the stochastic linearization in \eqref{stochastic linearization}, the bias induced by flatness incompatibility, as revealed in \textbf{Lemmas 1} and \textbf{2}, can lead to inconsistent flat regions of FedSAM at the cost of global flatness, hence degrading training performance when data heterogeneity is severe~\cite{pmlr-v235-lee24aa,qu2022generalized,10269141,pmlr-v202-sun23h,pmlr-v235-fan24c}.
To address flatness incompatibility and accelerate training, we design variance suppression operations on
(i) flatness search at the devices, (ii) local update at the devices, and (iii) global update at the PS.

\textbf{Variance Suppression on Local Flatness Search.} 
To enhance the consistency of local flatness under data heterogeneity and friendly adversary, we seek a stable stochastic linearization. In each training round $t$, we design the local perturbation rule to acquire the perturbed parameter $\tilde{\theta}_i^{t,k}$ as
\begin{align} \label{vs flatness search}
    \tilde{\theta}_i^{t,k} = \theta_i^{t,k} + \rho \frac{h_i^{t,k}}{ \| h_i^{t,k} \|} \ \text{with} \
    h_i^{t,k} = (1-\gamma_l)h^{t} + \gamma_l g_i^{t,k},
\end{align}
where $ g_i^{t,k} =  \nabla F_i(\theta_i^{t,k};\phi_i^{t,k}) $ is the stochastic gradient; 
$h_i^{t,k}$ is a convex combination of the PS-maintained adjusted direction $h^t$ and $g_i^{t,k}$, controlled by the local mixing coefficient $\gamma_l$.
This combination processes information from different devices' data to alleviate the impact of the friendly adversary.
A smaller $\gamma_l$ suppresses heterogeneity-induced device drift and reduces the local gradient variance on $ \frac{h_i^{t,k}}{ \| h_i^{t,k} \|}$.
When $\gamma_l \!\rightarrow\! 1$, $h_i^{t,k}$ reduces to $g_i^{t,k}$, and $\rho \frac{h_i^{t,k}}{ \| h_i^{t,k} \|}$ is determined by $g_i^{t,k}$ and fully exposed to both data heterogeneity and friendly adversary. In this case, the devices construct $\epsilon_i^{t,k}$ in inconsistent directions, and the perturbed gradients are evaluated at mismatched local points before aggregation at the PS. 
This reproduces the perturbation inconsistency in FedSAM identified in Section~\ref{subsec:flat-incompat}, which causes devices to generate mutually incompatible local flat regions.

\textbf{Variance Suppression on Local Update.}
Unlike centralized training~\cite{pmlr-v162-andriushchenko22a,pmlr-v139-kwon21b,Zhang_2023_CVPR,li2024enhancing}, the local gradient divergence induced by data heterogeneity in FL can severely impede SAM optimization. To address this, we design the local update with a variance-suppressed mixed direction $u_i^{t,k}$:
\begin{align} \label{vs local update}
\theta_i^{t,k+1}
\!=\! \theta_i^{t,k} \!-\! \eta_l u_i^{t,k} \ \text{with} \ u_i^{t,k} \!=\! (1 \!-\! \gamma_l) h^{t} \!+\! \gamma_l \tilde{g}_i^{t,k},
\end{align}
where $\tilde{g}_i^{t,k} = \nabla F_i(\tilde{\theta}_i^{t,k}; \phi_i^{t,k})$ is the perturbed local gradient, and $\eta_l$ denotes the local learning rate.

As proved in Appendix~\ref{app:local-var}, conditioning on the $\sigma$-algebra $\mathcal{F}^{t,k}$, it yields $\mathrm{Var}[u_i^{t,k}\!\mid\!\mathcal{F}^{t,k}] \!=\! \gamma_l^2\cdot\mathrm{Var}[\tilde g_i^{t,k}\!\mid\!\mathcal{F}^{t,k}]$, indicating $\gamma_l\!<\!1$ suppresses the stochastic variance of the mixed update direction $u_i^{t,k}$.
Using the same $\gamma_l$ in \eqref{vs flatness search} and \eqref{vs local update} makes $h_i^{t,k}$ and $u_i^{t,k}$ share the same adjusted direction $h^t$. This reduces the mismatch between the direction used to construct $\tilde{\theta}_i^{t,k}$ and the subsequent update direction, thereby alleviating the additional bias in $u_i^{t,k}$ caused by evaluating $\tilde{g}_i^{t,k}$ at $\tilde{\theta}_i^{t,k}$ generated from a mismatched search direction.
Consequently, the local flatness search in \eqref{vs flatness search} and the local update in \eqref{vs local update} share the same $h^t$, which improves consistency between the perturbation and update directions per device. Moreover, the $\gamma_l$-weighted local gradients $g_i^{t,k}$ and $\tilde{g}_i^{t,k}$ retain device-specific loss information and preserve the responsiveness of local flatness search and update to local data.

\textbf{Variance Suppression on Global Update.}
After local training, each selected device sends $\theta_i^{t}=\theta_i^{t,K}$ to the PS, which aggregates and scales the local updates to form $ g^{t+1} \!=\! \frac{1}{\eta_l K}(\theta^t \!-\! \frac{1}{S}\sum_{i \in \mathcal{S}^t}\theta_i^{t} ) \!=\! \frac{1}{SK}\sum_{i \in \mathcal{S}^t}\sum_{k=0}^{K-1} u_i^{t,k}$.
Here, $ \theta^t \!-\! \frac{1}{S}\sum_{i \in \mathcal{S}^t}\theta_i^{t} $, is divided by $\eta_l K$ to ensure $g^{t+1}$ directly comparable with $h^t$ in scale. Moreover, averaging over $S$ devices reduces stochastic fluctuation in $g^{t+1}$.

The global variance-suppressed direction $h^{t+1}$ is then updated by exponentially moving average (EMA), consistent with the variance suppression used in \eqref{vs flatness search} and \eqref{vs local update}:
\begin{align} \label{vs global update}
    \theta^{t+1} = \theta^t \!-\! \eta_g h^{t+1} \ \text{with} \ h^{t+1} = (1-\gamma_g)h^{t} +\gamma_g g^{t+1},
\end{align}
where $\eta_g$ and $\gamma_g$ denote the global learning rate and the global mixing coefficient, respectively. 
 Since $h^{t+1}$ is sent to the selected devices in the next training round $t+1$ and used in \eqref{vs flatness search} and \eqref{vs local update}, the tradeoff between gradient smoothing and local adaptation is controlled by $\gamma_g\gamma_l$: A smaller value places more weight on $h^t$ and less on the current averaged perturbed local gradients.

These variance-suppression operations cope with $\delta_i(\theta)$ and $\zeta_{i,\phi_i}(\theta)$ identified in \textbf{Lemmas~1} and~\textbf{2} at different stages of FedVSSAM. The local flatness search design in \eqref{vs flatness search} reduces the sensitivity of the perturbation direction to data heterogeneity and the local gradient deviation $\zeta_{i,\phi_i}(\theta)$ by replacing the local stochastic direction with a mixed direction anchored by $h^t$. The local update design in \eqref{vs local update} suppresses the stochastic variability associated with the batch gradient variance and, by sharing the same $h^t$ with \eqref{vs flatness search}, mitigates the propagation of heterogeneity-induced device drift from perturbation construction to local descent. The global update in \eqref{vs global update} further smooths the accumulated local updates, reducing the residual effects of both device drift and batch gradient variability in the aggregated direction. Hence, these operations jointly prevent the two sources of flatness incompatibility from being amplified across local flatness search, local update, and global update.

\section{Convergence Guarantee of FedVSSAM Under Non-convex Setting} \label{convergence sec}

This section analyzes the convergence and variance suppression of FedVSSAM. 

\subsection{Convergence Theorem}
Considering non-convex settings, under \textbf{Assumptions 1--3}, we establish the following theorem: 
\begin{theorem} \label{convergence of fedvssam}
Suppose that $F_i, \forall i \in \mathcal{N}$ is non-convex. Let the learning rates satisfy $\eta_l \leq \frac{1}{\sqrt{2}\gamma_l KL}$ and $\eta_g \leq \min \{ \frac{1}{24L},\frac{\gamma_l\gamma_g}{6L} \}$. Given the optimal global model $\theta^*$, the convergence upper bound of FedVSSAM with $T$ training rounds is given as
\begin{align} \label{convergence of fedvssam eq}
    \frac{1}{T} \sum_{t=0}^{T-1}  \mathbb{E} \Big[\big\|\nabla  F\big(\tilde{\theta}^t\big)\big\|^2\Big] &  \leq  
    \frac{(125\gamma_l \gamma_g \!+\! 88\eta_g L)\Delta}{20\eta_g \gamma_l \gamma_g T} \!+\! \frac{11}{5\gamma_l\gamma_g } a_{12} + \frac{31 e^2}{5} ( \gamma_l \eta_l K L)^2 G^0,
\end{align}
where $\Delta = F(\tilde{\theta}^0)-F^*$, $G^0 = \frac{1}{N} \sum_{i \in \mathcal{N}}\|\nabla F_i(\tilde{\theta}^0)\|^2$, $e$ denotes the Euler's number, and the expression of $a_{12}$ is given in \eqref{app:eq:a12}. The full proof can be found in Appendix~\ref{app:proof-theorem1}.
\end{theorem}
As revealed in \textbf{Theorem~\ref{convergence of fedvssam}}, with the variance suppression operations designed in \eqref{vs flatness search}--\eqref{vs global update}, FedVSSAM enables the convergence of non-convex models under data heterogeneity and partial device participation. The first term on the RHS of \eqref{convergence of fedvssam eq} vanishes as $T \to \infty$, while the non-vanishing term $\frac{11}{5\gamma_l\gamma_g } a_{12}$ characterizes the effects of data heterogeneity ($\sigma_{g,1}^2$), friendly adversary ($\sigma_{l,1}^2$), and partial device participation ($\frac{N-S}{S(N-1)}$).
Moreover, the choices of the mixing coefficients $\gamma_l$ and $\gamma_g$ exhibit a trade-off for FedVSSAM. A smaller $\gamma_l \gamma_g$ shrinks $\frac{11}{5\gamma_l\gamma_g } a_{12} + \frac{31 e^2}{5} ( \gamma_l \eta_l K L)^2 G^0$, leading to a decreased non-vanishing error, while it can enlarge $\frac{(125\gamma_l \gamma_g \!+\! 88\eta_g L)}{20\eta_g \gamma_l \gamma_g }$ and retard convergence. As the proportion of participating devices $\frac{S}{N}$ increases, the non-vanishing error $\frac{11}{5\gamma_l\gamma_g } a_{12}$ decreases, indicating that more device participation can improve convergence.
\begin{corollary} \label{convergence corollary}
Define $\Sigma = \frac{\sigma_{l,1}^2}{N K} + \frac{N-S}{S(N-1)}\sigma^2$ with $\sigma^2 = \sigma_{l,1}^2+\sigma_{g,1}^2$ and $\beta=\gamma_l \gamma_g$. Set $\beta=\min\!\left\{\,1,~\sqrt{\frac{L \Delta}{\Sigma T}}~\right\}$, the global and local learning rates $\eta_g \leq \min\!\left\{\frac{1}{24L},\frac{\beta}{6L}\right\}$ and $\eta_l \leq \frac{1}{\sqrt{2} \gamma_l K L}$, $ \eta_l K L \lesssim \min\!\left\{1,\big(\frac{L\Delta}{G^0\,\beta^3 T}\big)^{0.5},\frac{1}{(\beta S)^{0.5}},\frac{1}{(\beta^3 S K)^{0.25}}\right\}$, and $\rho \lesssim \frac{1}{L\,(S K T)^{0.25}}$, FedVSSAM guarantees
\begin{align} \label{convergence of fedvssam corollary}
    \frac{1}{T} \sum_{t=0}^{T-1}  \mathbb{E} \Big[\big\|\nabla  F\big(\tilde{\theta}^t\big)\big\|^2\Big] \! \leq \! \frac{L \Delta}{T} \!+\! \sqrt{ \frac{L \Delta \sigma_{l,1}^2}{NKT} } \!+\! \sqrt{\frac{L\Delta \sigma^2}{ST}} .
\end{align}
\end{corollary}
It is indicated in \textbf{Corollary~\ref{convergence corollary}} that under suitable choices of $\eta_l$ and $\eta_g$ and $\gamma_l$ and $\gamma_g$, FedVSSAM achieves a linear speedup with the number of participating devices, $S$, and the number of local training iterations, $K$. 
Without requiring devices to transmit additional auxiliary vectors, such as control or dual variables, FedVSSAM guarantees $\mathcal{O}(\frac{1}{T}+\frac{1}{\sqrt{NKT}}+\frac{1}{\sqrt{ST}})$, incurring no additional convergence penalty that scales with $K$. A further discussion of the convergence rates and communication requirements are in Appendix~\ref{app:conv-comm-comparison}.

\subsection{Discussion}

FedVSSAM can be regarded as an extension of FedSAM. When the mixing coefficients $\gamma_l= \gamma_g = 1$ and the FedAvg-style aggregation scheme are employed, FedVSSAM degenerates to the vanilla FedSAM.
With the variance suppression designs on flatness search, local and global updates in \eqref{vs flatness search}, \eqref{vs local update}, and \eqref{vs global update}, respectively, FedVSSAM alleviates the impacts of data heterogeneity and friendly adversary, addressing flatness incompatibility.
To further expose the effects of the proposed variance suppression designs, i.e., \eqref{vs flatness search}, \eqref{vs local update}, and \eqref{vs global update}, we establish the following theorem to quantify how accurately the adjusted direction $h^t$ tracks the target global gradient.
\begin{theorem} \label{the good of variance suppression}
Under \textbf{Assumptions 1--3}, following the definitions and settings of $\Sigma$, $\beta$, $\eta_g$, $\eta_l$, and $\eta_l K L$ in \textbf{Corollary~\ref{convergence corollary}}, FedVSSAM guarantees the MSE between the adjusted gradient, $h^t$, and the global gradient, $\nabla F(\theta^t)$, is upper bounded by 
\begin{align} \label{the good of variance suppression eq}
    \frac{1}{T}\!\!\sum_{t=1}^{T}\mathbb{E}\!\left[\| h^t \!\!-\!\! \nabla F(\theta^t) \|^2\right] \! \lesssim \!
\frac{L\Delta}{\beta T} \!+\! \beta \Sigma \!+\!
\sqrt{\frac{L\Delta \Sigma}{T}} \!+\! \frac{\Sigma}{\sqrt{T}} ,
\end{align}
where $\Delta = F(\tilde{\theta}^0)-F^*$. The full proof can be found in Appendix~\ref{proof-theorem2 subsec}.
\end{theorem}
\textbf{Theorem~\ref{the good of variance suppression}} quantifies the effects of variance suppression~\eqref{vs flatness search}, \eqref{vs local update}, and \eqref{vs global update} and the role of the mixing coefficients $\gamma_l$ and $\gamma_g$. 
The term $\tfrac{L\Delta}{\beta T}$ on the RHS of \eqref{the good of variance suppression eq} vanishes increasingly fast with the rise of $\beta$, whereas the residual variance $\beta\Sigma$ is reduced due to a smaller $\beta$. 
This reaffirms the role of $\beta$ to balance the vanishing term $\frac{L \Delta}{\beta T}$ and the residual variance $\beta \Sigma$: Increasing $\beta$ can accelerate early-stage convergence but enlarges the residual variance, whereas decreasing $\beta$ suppresses the residual variance at the cost of a slower reduction of $\frac{L \Delta}{\beta T}$.
A further discussion can be found in Appendix~\ref{discussion-theorem2 subsec}.

\section{Experiment}
In this section, we empirically evaluate the proposed FedVSSAM in enhancing generalization and accelerating training. All experiments are run on NVIDIA 4090 GPUs.

\subsection{Experimental Setup}

We compare FedVSSAM with the vanilla FedAvg~\cite{fedavg}, and SAM-based FL algorithms, including FedSAM~\cite{caldarola2022improving}, MoFedSAM~\cite{qu2022generalized}, FedGAMMA~\cite{10269141}, FedSMOO~\cite{pmlr-v202-sun23h}, and FedLESAM~\cite{pmlr-v235-fan24c}. We also consider the representative state-of-the-art heterogeneity mitigation and customized aggregation FL algorithms, including FedProx~\cite{fedprox}, SCAFFOLD~\cite{scaffold}, FedCM~\cite{xu2021fedcm}, FedDyn~\cite{feddyn}, FedAdam~\cite{reddi2021adaptive}, and FedACG~\cite{kim2024communication}.

Consider 100/200-device FL systems with 10\% and 5\% of devices selected UAR per round, respectively. 
CIFAR-10 and CIFAR-100 datasets~\cite{krizhevsky2009learning} are considered with the ResNet-18 model~\cite{he2016deep} and the WRN-28-4 model \cite{zagoruyko2016wide}, respectively.
To characterize data heterogeneity, we utilize both the Dirichlet distribution with parameter $\alpha$, and the Pathological distribution with parameter $c$~\cite{pmlr-v235-lee24aa,qu2022generalized,10269141,pmlr-v202-sun23h,pmlr-v235-fan24c}.
The corresponding parameters $\alpha$ and $c$ are selected as $\{0.1,0.5\}$ and $\{3,6\}$ for CIFAR-10, and $\{0.1,0.5\}$ and $\{10,20\}$ for CIFAR-100, to account for moderate and severe data heterogeneity, respectively.
All algorithms are trained for 800 rounds; the learning rate grids, batch sizes, local epochs, perturbation radius, and specific algorithm parameters are summarized in Appendix~\ref{app:algorithm-parameters}. We also conduct text topic classification on DBpedia-14 ~\cite{zhang2015character} with a TextCNN~\cite{kim2014convolutional} to demonstrate the performance of FedVSSAM in text classification, see Appendix~\ref{app:dbpedia-text}.

\subsection{Evaluation of FedVSSAM}

\begin{table*}
\centering
\caption{Test accuracy of FedVSSAM and baselines on the CIFAR-10/100 dataset after 800 training rounds. The device sampling setups are 10\%-100 devices (upper part) and 5\%-200 devices (lower part), respectively. Acc. means the final accuracy, and Rd. represents the number of training rounds required for an algorithm to first reach a certain accuracy level, which is 75\% and 40\% for CIFAR-10 and CIFAR-100, respectively.
A dash (--) indicates that the corresponding target accuracy was not reached within the training budget under that setting.
}
\label{acc table}
\begingroup
\renewcommand{\arraystretch}{1.25}
\resizebox{\textwidth}{!}{%
\begin{tabular}{l|cccccccc|cccccccc} 
\hline
\multicolumn{1}{c|}{\multirow{4}{*}{\begin{tabular}[c]{@{}c@{}}Dataset \\Partition\\~Coefficient\end{tabular}}} & \multicolumn{8}{c|}{CIFAR-10}                                                                                                         & \multicolumn{8}{c}{CIFAR-100}                                                                                                           \\ 
\cline{2-17}
\multicolumn{1}{c|}{}                                                                                           & \multicolumn{4}{c}{Dirichlet}                                           & \multicolumn{4}{c|}{Pathological}                          & \multicolumn{4}{c}{Dirichlet}                                           & \multicolumn{4}{c}{Pathological}                             \\ 
\cline{2-17}
\multicolumn{1}{c|}{}                                                                                           & \multicolumn{2}{c}{$\alpha{=}0.5$} & \multicolumn{2}{c}{$\alpha{=}0.1$} & \multicolumn{2}{c}{$c{=}6$} & \multicolumn{2}{c|}{$c{=}3$} & \multicolumn{2}{c}{$\alpha{=}0.5$} & \multicolumn{2}{c}{$\alpha{=}0.1$} & \multicolumn{2}{c}{$c{=}20$} & \multicolumn{2}{c}{$c{=}10$}  \\ 
\cline{2-17}
\multicolumn{1}{c|}{}                                                                                           & Acc.           & Rd.               & Acc.           & Rd.               & Acc.           & Rd.        & Acc.           & Rd.         & Acc.           & Rd.               & Acc.           & Rd.               & Acc.           & Rd.         & Acc.           & Rd.          \\ 
\hline
FedAvg ~\cite{fedavg}                                                                                           & {$77.05_{\pm .14}$} & {315}               & {$72.57_{\pm .19}$} & {--}               & {$77.55_{\pm .24}$} & {299}        & {$74.63_{\pm .23}$} & {--}         & {$46.52_{\pm .23}$} & {313}               & {$36.76_{\pm .23}$} & {--}               & {$42.59_{\pm .15}$} & {486}         & {$36.67_{\pm .21}$} & {--}          \\
FedProx~\cite{fedprox}                                                                                          & {$76.82_{\pm .14}$} & {337}               & {$72.23_{\pm .16}$} & {--}               & {$77.81_{\pm .19}$} & {271}        & {$74.89_{\pm .20}$} & {--}         & {$46.80_{\pm .24}$} & {313}               & {$37.00_{\pm .17}$} & {--}               & {$41.60_{\pm .20}$} & {517}         & {$36.75_{\pm .19}$} & {--}          \\
FedAdam~\cite{reddi2021adaptive}                                                                                & {$78.35_{\pm .13}$} & {191}               & {$75.76_{\pm .24}$} & {741}               & {$78.13_{\pm .21}$} & {188}        & {$77.26_{\pm .21}$} & {440}         & {$54.80_{\pm .24}$} & {155}               & {$48.29_{\pm .28}$} & {276}               & {$50.38_{\pm .24}$} & {376}         & {$43.36_{\pm .24}$} & {637}          \\
SCAFFOLD~\cite{scaffold}                                                                                        & {$80.14_{\pm .16}$} & {170}               & {$75.81_{\pm .18}$} & {635}               & {$81.35_{\pm .15}$} & {136}        & {$78.72_{\pm .15}$} & {326}         & {$\mathbf{59.96}_{\pm .32}$} & {274}               & {$50.59_{\pm .28}$} & {450}               & {$\mathbf{55.78}_{\pm .33}$} & {368}         & {$47.05_{\pm .23}$} & {534}          \\
FedACG~\cite{kim2024communication}                                                                              & {$78.84_{\pm .18}$} & {\textbf{113}}               & {$76.85_{\pm .15}$} & {457}               & {$80.62_{\pm .15}$} & {\textbf{85}}        & {$79.02_{\pm .14}$} & {\textbf{199}}         & {$51.66_{\pm .20}$} & {\textbf{106}}               & {$46.30_{\pm .30}$} & {\textbf{161}}               & {$51.85_{\pm .30}$} & {\textbf{124}}         & {$46.18_{\pm .17}$} & {\textbf{232}}          \\
FedCM~\cite{xu2021fedcm}                                                                                        & {$76.23_{\pm .21}$} & {388}               & {$74.07_{\pm .21}$} & {--}               & {$77.48_{\pm .19}$} & {212}        & {$76.14_{\pm .19}$} & {574}         & {$48.51_{\pm .15}$} & {\textbf{145}}               & {$44.63_{\pm .23}$} & {\textbf{254}}               & {$49.71_{\pm .22}$} & {226}         & {$45.32_{\pm .27}$} & {343}          \\
FedDyn~\cite{feddyn}                                                                                            & {$76.80_{\pm .16}$} & {356}               & {$74.60_{\pm .17}$} & {--}               & {$77.19_{\pm .19}$} & {255}        & {$76.78_{\pm .17}$} & {446}         & {$50.84_{\pm .26}$} & {177}               & {$44.83_{\pm .25}$} & {307}               & {$50.40_{\pm .14}$} & {\textbf{209}}         & {$44.72_{\pm .28}$} & {377}          \\
FedSAM~\cite{caldarola2022improving}                                                                            & {$78.75_{\pm .14}$} & {266}               & {$74.06_{\pm .21}$} & {--}               & {$80.25_{\pm .13}$} & {240}        & {$76.07_{\pm .14}$} & {639}         & {$52.22_{\pm .17}$} & {439}               & {$43.50_{\pm .15}$} & {688}               & {$47.52_{\pm .17}$} & {530}        & {$39.84_{\pm .32}$} & {--}          \\
MoFedSAM~\cite{qu2022generalized}                                                                               & {$81.91_{\pm .16}$} & {199}               & {$\mathbf{78.94}_{\pm .14}$} & {385}               & {$\mathbf{82.94}_{\pm .16}$} & {178}        & {$81.15_{\pm .17}$} & {291}         & {$47.14_{\pm .18}$} & {576}               & {$49.69_{\pm .15}$} & {450}               & {$54.49_{\pm .29}$} & {360}         & {$45.60_{\pm .17}$} & {589}          \\
FedGAMMA~\cite{10269141}                                                                                        & {$82.36_{\pm .13}$} & {201}               & {$77.02_{\pm .15}$} & {\textbf{208}}               & {$82.74_{\pm .18}$} & {179}        & {$79.49_{\pm .19}$} & {393}         & {$51.39_{\pm .26}$} & {511}               & {$38.91_{\pm .24}$} & {--}               & {$45.29_{\pm .22}$} & {638}        & {$35.57_{\pm .14}$} & {--}          \\
FedSMOO~\cite{pmlr-v202-sun23h}                                                                                 & {$\mathbf{81.49}_{\pm .16}$} & {147}               & {$78.85_{\pm .22}$} & {358}               & {$82.87_{\pm .12}$} & {134}        & {$\mathbf{80.76}_{\pm .23}$} & {289}         & {$56.87_{\pm .23}$} & {279}               & {$\mathbf{49.12}_{\pm .30}$} & {396}               & {$55.04_{\pm .21}$} & {330}         & {$\mathbf{48.60}_{\pm .18}$} & {463}          \\
FedLESAM~\cite{pmlr-v235-fan24c}                                                                                & {$76.88_{\pm .17}$} & {306}               & {$72.31_{\pm .18}$} & {--}              & {$77.64_{\pm .23}$} & {240}        & {$75.49_{\pm .17}$} & {736}         & {$46.58_{\pm .15}$} & {308}               & {$35.85_{\pm .17}$} & {--}               & {$43.19_{\pm .32}$} & {458}         & {$36.33_{\pm .21}$} & {--}          \\
FedVSSAM(ours)                                                                                                  & {$\mathbf{83.56}_{\pm .19}$} & {\textbf{111}}               & {$\mathbf{79.57}_{\pm .14}$} & {\textbf{296}}               & {$\mathbf{83.57}_{\pm .22}$} & {\textbf{107}}        & {$\mathbf{81.11}_{\pm .23}$} & {\textbf{196}}         & {$\mathbf{58.78}_{\pm .30}$} & {237}               & {$\mathbf{51.22}_{\pm .20}$} & {328}               & {$\mathbf{56.87}_{\pm .19}$} & {258}         & {$\mathbf{50.27}_{\pm .18}$} & {\textbf{337}}          \\ 
\hline
FedAvg ~\cite{fedavg}                                                                                           & {$74.86_{\pm .27}$} & {--}               & {$70.30_{\pm .36}$} & {--}               & {$75.15_{\pm .21}$} & {789}        & {$72.74_{\pm .37}$} & {--}         & {$32.45_{\pm .38}$} & {--}               & {$28.29_{\pm .28}$} & {--}               & {$30.39_{\pm .21}$} & {--}         & {$27.74_{\pm .28}$} & {--}          \\
FedProx~\cite{fedprox}                                                                                          & {$74.27_{\pm .17}$} & {--}               & {$70.82_{\pm .23}$} & {--}               & {$74.65_{\pm .25}$} & {--}        & {$72.10_{\pm .31}$} & {--}         & {$31.56_{\pm .21}$} & {--}               & {$33.21_{\pm .18}$} & {--}               & {$29.92_{\pm .35}$} & {--}         & {$27.18_{\pm .21}$} & {--}         \\
FedAdam~\cite{reddi2021adaptive}                                                                                & {$75.05_{\pm .32}$} & {751}               & {$74.56_{\pm .28}$} & {--}               & {$75.57_{\pm .17}$} & {615}        & {$76.17_{\pm .20}$} & {485}         & {$45.96_{\pm .19}$} & {259}               & {$43.43_{\pm .20}$} & {494}               & {$39.65_{\pm .22}$} & {--}         & {$41.29_{\pm .21}$} & {720}          \\
SCAFFOLD~\cite{scaffold}                                                                                        & {$77.79_{\pm .35}$} & {299}               & {$74.19_{\pm .21}$} & {--}               & {$78.13_{\pm .16}$} & {264}       & {$76.18_{\pm .29}$} & {549}         & {$53.96_{\pm .24}$} & {277}              & {$47.85_{\pm .22}$} & {484}               & {$46.27_{\pm .20}$} & {455}         & {$45.63_{\pm .21}$} & {584}          \\
FedACG~\cite{kim2024communication}                                                                              & {$78.03_{\pm .35}$} & {\textbf{213}}               & {$75.64_{\pm .29}$} & {603}               & {$77.71_{\pm .25}$} & {258}        & {$77.68_{\pm .23}$} & {\textbf{331}}         & {$45.77_{\pm .31}$} & {\textbf{152}}               & {$44.48_{\pm .29}$} & {\textbf{246}}               & {$45.73_{\pm .33}$} & {\textbf{240}}         & {$40.31_{\pm .31}$} & {670}          \\
FedCM~\cite{xu2021fedcm}                                                                                        & {$72.88_{\pm .22}$} & {--}               & {$72.01_{\pm .17}$} & {--}               & {$73.93_{\pm .37}$} & {--}        & {$74.25_{\pm .20}$} & {--}         & {$45.77_{\pm .26}$} & {183}               & {$47.93_{\pm .31}$} & {\textbf{269}}               & {$43.11_{\pm .26}$} & {379}         & {$42.81_{\pm .20}$} & {\textbf{467}}          \\
FedDyn~\cite{feddyn}                                                                                            & {$72.53_{\pm .17}$} & {--}               & {$71.23_{\pm .30}$} & {--}               & {$72.94_{\pm .16}$} & {--}        & {$73.48_{\pm .18}$} & {--}         & {$35.53_{\pm .34}$} & {--}               & {$32.73_{\pm .20}$} & {--}               & {$34.60_{\pm .24}$} & {--}         & {$30.78_{\pm .33}$} & {--}          \\
FedSAM~\cite{caldarola2022improving}                                                                            & {$76.17_{\pm .35}$} & {581}               & {$71.84_{\pm .33}$} & {--}               & {$77.38_{\pm .20}$} & {394}        & {$74.19_{\pm .28}$} & {--}         & {$46.99_{\pm .22}$} & {448}               & {$39.63_{\pm .36}$} & {--}               & {$40.75_{\pm .23}$} & {723}         & {$35.84_{\pm .27}$} & {--}          \\
MoFedSAM~\cite{qu2022generalized}                                                                               & {$81.93_{\pm .26}$} & {216}               & {$79.00_{\pm .32}$} & {\textbf{404}}               & {$82.65_{\pm .25}$} & {\textbf{178}}        & {$80.13_{\pm .21}$} & {332}         & {$47.12_{\pm .33}$} & {580}               & {$48.47_{\pm .37}$} & {513}               & {$\mathbf{55.16}_{\pm .30}$} & {349}         & {$\mathbf{47.40}_{\pm .24}$} & {572}          \\
FedGAMMA~\cite{10269141}                                                                                        & {$81.17_{\pm .31}$} & {218}               & {$76.99_{\pm .25}$} & {574}               & {$81.00_{\pm .37}$} & {203}        & {$78.85_{\pm .37}$} & {433}         & {$52.86_{\pm .31}$} & {466}               & {$40.09_{\pm .30}$} & {799}               & {$48.38_{\pm .35}$} & {584}         & {$36.12_{\pm .24}$} & {--}          \\
FedSMOO~\cite{pmlr-v202-sun23h}                                                                                 & {$\mathbf{78.28}_{\pm .30}$} & {246}               & {$\mathbf{77.31}_{\pm .27}$} & {512}               & {$79.70_{\pm .31}$} & {210}        & {$78.48_{\pm .20}$} & {393}         & {$\mathbf{54.02}_{\pm .31}$} & {289}               & {$\mathbf{50.13}_{\pm .32}$} & {395}               & {$49.96_{\pm .23}$} & {333}         & {$46.84_{\pm .20}$} & {472}          \\
FedLESAM~\cite{pmlr-v235-fan24c}                                                                                & {$74.61_{\pm .37}$} & {--}               & {$69.89_{\pm .23}$} & {--}               & {$75.14_{\pm .37}$} & {708}        & {$72.69_{\pm .37}$} & {--}         & {$31.97_{\pm .25}$} & {--}               & {$34.75_{\pm .30}$} & {--}               & {$30.11_{\pm .35}$} & {--}         & {$28.24_{\pm .37}$} & {--}          \\
FedVSSAM(ours)                                                                                                  & {$\mathbf{83.14}_{\pm .30}$} & {\textbf{145}}               & {$\mathbf{79.42}_{\pm .23}$} & {\textbf{281}}               & {$\mathbf{83.52}_{\pm .32}$} & {\textbf{131}}        & {$\mathbf{80.90}_{\pm .29}$} & {\textbf{236}}         & {$\mathbf{57.53}_{\pm .26}$} & {248}               & {$\mathbf{50.45}_{\pm .32}$} & {376}               & {$\mathbf{55.94}_{\pm .26}$} & {\textbf{316}}         & {$\mathbf{50.17}_{\pm .19}$} & {\textbf{398}}          \\
\hline
\end{tabular}%
}
\endgroup
\end{table*}

As illustrated in Table~\ref{acc table}, FedVSSAM consistently outperforms the baselines across datasets, models, heterogeneity levels, and sampling settings, validating its effectiveness in addressing flatness incompatibility and enhancing generalization; the convergence curve analysis is provided in Appendix~\ref{app:convergence-curves}.
Compared with the second-best algorithm, FedVSSAM improves accuracy by 1.4\% and 1.7\% with $\alpha=0.5$ and 100 devices on CIFAR-10 and CIFAR-100, respectively. When data heterogeneity becomes intense (e.g., $\alpha =0.1$), FedVSSAM still achieves considerable performance enhancement over the baselines, attributed to its efficient implementation of a more globally consistent flatness search, enabling the model to achieve enhanced generalization under non-IID data.

Table~\ref{acc table} also demonstrates that FedVSSAM exhibits a significant advantage in communication efficiency as well. For instance, FedVSSAM takes 111 rounds to reach 75\% accuracy for CIFAR-10 with 100 devices and $\alpha = 0.5$, which requires $64.8\%$ fewer rounds than FedAvg.
Variance suppression operations on both devices and the PS side enable the designed adjusted direction $h^{t+1}$ to rapidly align with the ideal $\nabla F(\theta^{t+1})$, thereby achieving effective convergence acceleration.

\textbf{Visualization Results.} We conduct t-SNE visualization~\cite{van2008visualizing} of feature embeddings to provide qualitative evidence for the representation quality and generalization advantage of FedVSSAM; as detailed in Appendix~\ref{app:tsne-visualization}.

\begin{figure}[htbp]
	\centering
	\begin{minipage}{0.49\textwidth}
		\centering
		\subfloat[$\alpha=0.5$.]{\includegraphics[width=.48\linewidth]{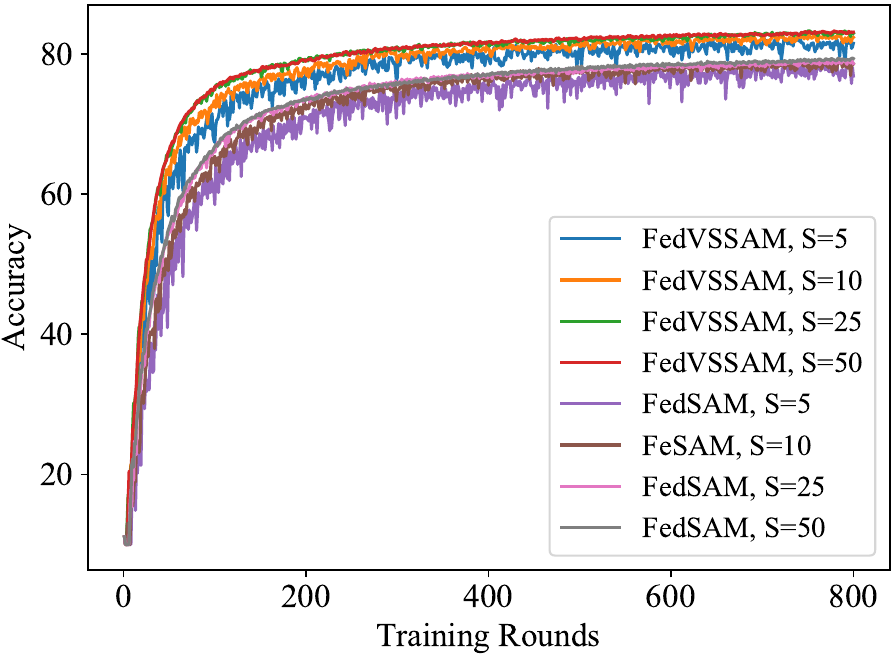}
			\label{fig_ue_1}}
		\subfloat[$\alpha=0.1$.]{\includegraphics[width=.48\linewidth]{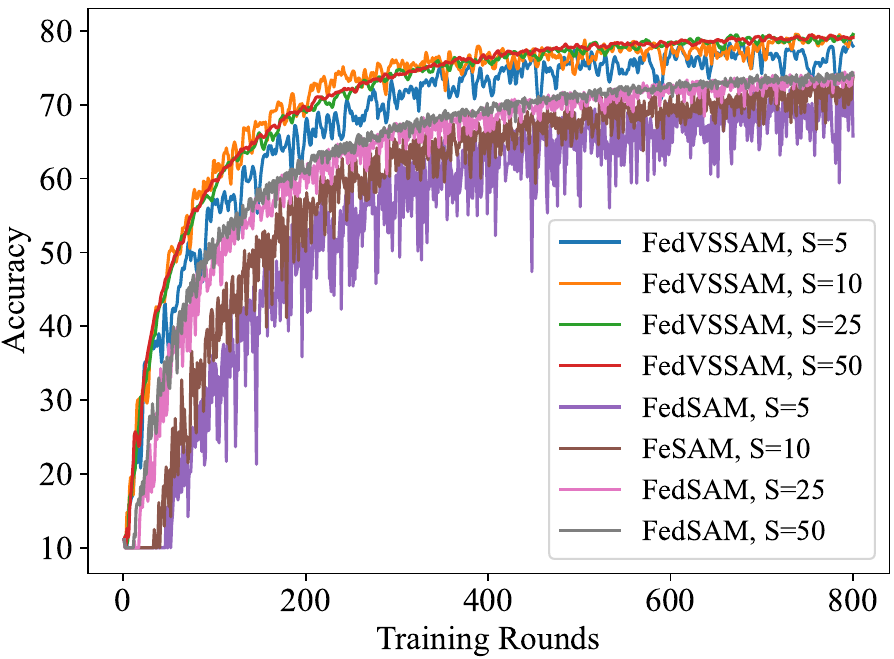}
			\label{fig_ue_2}}
		\caption{Accuracy with different numbers of participating devices on CIFAR-10.}
		\label{fig_ue}
	\end{minipage}\hfill
	\begin{minipage}{0.49\textwidth}
		\centering
		\subfloat[$\alpha=0.5$.]{\includegraphics[width=.48\linewidth]{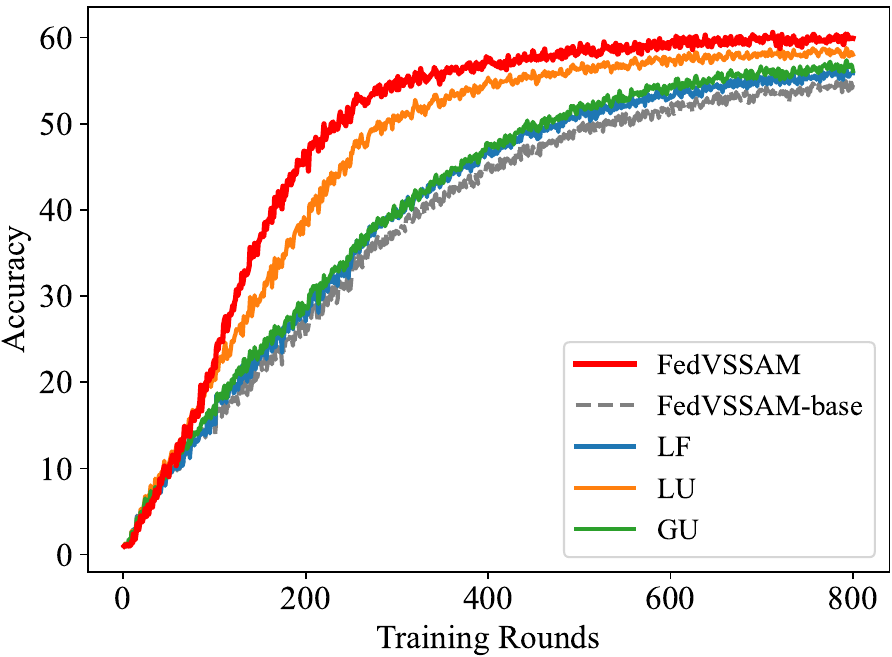}
			\label{fig_simplified_1}}
		\subfloat[$\alpha=0.1$.]{\includegraphics[width=.48\linewidth]{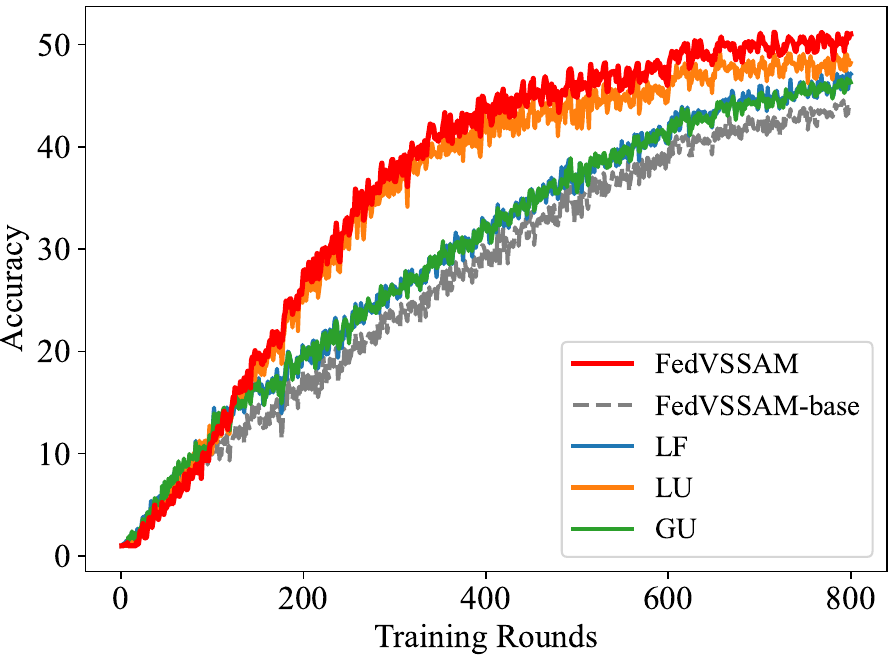}
			\label{fig_simplified_2}}
		\caption{Ablation studies on the variance suppression operations in FedVSSAM.}
		\label{fig_simp}
	\end{minipage}
\end{figure}

\textbf{Effect of Device Sampling.}
Fig.~\ref{fig_ue} evaluates FedVSSAM under different numbers of participating devices, where a UAR selection of $5\%$, $10\%$, $25\%$, and $50\%$ out of $N=100$ devices is considered.
Clearly, more device participation leads to faster convergence. 
A larger number of selected devices $S$ contributes to a more stable adjusted direction $h^{t}, \forall t$ in \eqref{vs global update} and a tighter bound on the RHS of \eqref{convergence of fedvssam corollary} and helps better guide local flatness search and gradient modifications.

\textbf{Global Direction Tracking and Flatness Incompatibility Suppression.}
We examine the global direction tracking and flatness incompatibility suppression of FedVSSAM.
The results provide additional illustration for the role of $h^t$ in variance suppression; see Appendix~\ref{app:flatness-incompatibility-probe}.

\textbf{Ablation Study of Variance Suppression Operations.}
In Fig.~\ref{fig_simp}, we conduct ablation studies on the variance suppression operations in FedVSSAM. By setting $\gamma_l=\gamma_g=1$ as the baseline of FedVSSAM, i.e., FedVSSAM-base, we individually activate the variance suppression of local flatness search~\eqref{vs flatness search}, local update~\eqref{vs local update}, and global update~\eqref{vs global update}, named LF, LU, and GU, respectively, to examine their respective gains on convergence. 
It is observed that all LF, LU, and GU achieve performance gains compared with FedVSSAM-base, indicating the effectiveness of each designed variance suppression operation.
Meanwhile, detailed sensitivity results on $\gamma_l$, $\gamma_g$, $K$, and $\rho$ in Appendix~\ref{app:sensitivity} show the robustness of FedVSSAM.

\section{Conclusion}

In this paper, we proposed FedVSSAM to address flatness incompatibility and expedite model training in SAM-based FL with designed variance suppression operations consistently in local flatness search, local update, and global aggregation. 
By identifying data heterogeneity and friendly adversary as two sources of flatness incompatibility, FedVSSAM provided a unified mechanism to stabilize perturbation construction, local updates, and global updates. We established non-convex convergence under data heterogeneity and partial device participation, and proved how the designed variance suppression operations control the MSE between the variance-suppressed adjusted direction and the global gradient. Numerical results further demonstrated that FedVSSAM achieves better convergence speed and final accuracy over both FL and SAM-based FL baselines in most of the cases.
The current study focuses on standard smoothness, bounded variance, and UAR device participation assumptions with simulation results on vision and text tasks. Extending FedVSSAM to broader model families, real-world hardware deployment, and integration with privacy mechanisms remains important future work.

\bibliographystyle{plainnat}
\bibliography{references}

\newpage
\appendix

\section{Additional Related Work}\label{app:additional-related-work}

\textbf{Heterogeneous FL.}
FL enables collaborative training without sharing personal raw data, yet the PS can only coordinate training through uploaded model updates from the devices, inducing non-IID data distributions as central challenges~\cite{fedavg,bonawitz2019towards,kairouz2021advances,XIAO2024}.
To tackle data heterogeneity, conventional studies typically modify local objectives, device updates, or model aggregation. FedProx uses a proximal term to restrict local drift, while SCAFFOLD corrects device drift with designed control variates~\cite{fedprox,scaffold}. FedNova adjusts heterogeneous local progress~\cite{fednova} with normalized aggregation coefficients. FedDyn introduces dynamic regularization techniques to improve model training~\cite{feddyn}.
Momentum-based methods improve stability from another perspective: FedCM adds device-level momentum, and recent theory shows that momentum can benefit non-IID FL~\cite{xu2021fedcm,cheng2024momentum}. PS-side adaptive optimization and the modified local gradients further improve training under heterogeneous updates~\cite{kingma2014adam,reddi2021adaptive,kim2024communication}.
Another line of heterogeneous FL shifts the training objective from one universal global model to device-aware objectives.
FedMA constructs the global model layer by layer by matching and averaging hidden elements with similar feature-extraction signatures~\cite{wang2020fedma}. FedBN keeps local batch normalization statistics, and personalized representation shares only part of the model~\cite{li2021fedbn,pmlr-v139-collins21a}. 
Ditto and other personalized FL also show that a single averaged model may be insufficient for all devices~\cite{pmlr-v139-li21h,dinh2020personalized,fallah2020personalized}. 
These studies mainly reduce device drifts or personalization mismatch, but do not explicitly target FL generalization, which is also crucial in FL because the global model is trained from partial and non-IID local data and is expected to serve heterogeneous or unseen device distributions~\cite{kairouz2021advances,pmlr-v97-mohri19a,caldarola2022improving}. 

\textbf{Sharpness-aware optimization.}
The connection between flat minima and generalization has been studied before SAM. Early flat-minima arguments suggested that broader basins tend to generalize better, and later empirical studies observed that models trained with large batches often converge to sharper minima with worse test performance~\cite{hochreiter1997flat,keskar2017on}.
 PAC-Bayesian and empirical studies further examine which sharpness-related quantities correlate with generalization in modern neural networks~\cite{dziugaite2017computing,jiang2019fantastic}. SAM converts this geometric intuition into a minimax objective by updating the model using gradients evaluated at an adversarially perturbed point~\cite{foret2021sharpnessaware}.
Subsequent work has refined the understanding and implementation of this perturbation. Analysis of SAM investigates when minimizing the perturbed loss reduces sharpness and how the perturbation affects the training trajectory~\cite{pmlr-v162-andriushchenko22a,wen2022how}. ASAM adapts the sharpness measure to scale-invariant parameterizations, and Fisher SAM interprets the perturbation through information geometry~\cite{pmlr-v139-kwon21b,pmlr-v162-kim22f}. Other methods adjust the geometry or cost of the SAM step: surrogate gap minimization changes the objective gap, efficient SAM reduces computation, and LookSAM improves scalability~\cite{zhuang2022surrogate,du2022efficient,liu2022looksam}. Gradient norm-aware minimization pursues first-order flatness, and Riemannian SAM extends the idea to manifold optimization~\cite{Zhang_2023_CVPR,yun2023riemannian}. Friendly SAM and VaSSO are closer to our motivation by studying when adversarial perturbations may become insufficiently adversarial~\cite{Li_2024_CVPR,li2024enhancing}.
FedVSSAM differs from centralized SAM optimizers since it targets distributed objectives. 
Centralized SAM variants usually train on batches sampled from a single data distribution, so different batches are treated as noisy estimates of the same objective. 
In SAM-based FL, however, this assumption no longer holds: each device owns its local objective under non-IID data, and the SAM perturbation constructed from one local batch may not remain adversarial for another local batch, which needs FedVSSAM to address both data heterogeneity and friendly adversary.

\textbf{SAM-based FL.}
FedSAM first brought SAM into FedAvg to improve generalization under decentralized training~\cite{caldarola2022improving}. MoFedSAM generalizes the framework and analyzes FedSAM under data heterogeneity~\cite{qu2022generalized}. 
FedSpeed pursues larger local intervals and improved communication efficiency by combining SAM with proximal control~\cite{sun2023fedspeed}. FedSMOO combines FedDyn and dynamic regularization to approach a smoother landscape~\cite{pmlr-v202-sun23h}. 
Differentially private FL has also been connected to flat landscapes, showing that sharpness-aware design can be useful when privacy noise is present~\cite{Shi_2023_CVPR,9347706}.

Other studies try to align perturbations with the global objective. FedGAMMA introduces global sharpness-aware minimization, and FedGF rethinks flat minima search by emphasizing local--global flatness discrepancy~\cite{10269141,pmlr-v235-lee24aa}. FedLESAM estimates global perturbations locally to avoid relying only on local perturbations~\cite{pmlr-v235-fan24c}. FedGloSS moves the sharpness search toward communication-efficient global sharpness, and FedGMT uses the global model trajectory to measure sharpness~\cite{Caldarola_2025_CVPR,pmlr-v267-li25bd}. FedWMSAM combined weighted momentum with dynamic SAM to address local--global curvature misalignment and momentum-echo oscillation~\cite{li2025fedwmsam}. 
Recent extensions also use synthetic data to guide SAM optimization under gradient compression~\cite{gu2026fedsynsam}, and study decentralized SAM-based FL settings~\cite{hu2026dfedlsam}.

FedVSSAM follows the global flatness view but uses a different design principle. Existing methods often improve or approximate the perturbation or sharpness through corrected local updates, PS-side sharpness, global trajectories, or momentum scheduling. 
FedVSSAM instead starts from the sources of flatness incompatibility, namely data heterogeneity and friendly adversary, and uses one adjusted direction to anchor local flatness search, local update, and global update. 
This design is reflected in both the algorithm and theory: the PS-maintained direction $h^t$ is reused across these stages, and its tracking error to the global gradient is effectively controlled through the MSE bound. 
Consequently, FedVSSAM mitigates flatness incompatibility without requiring devices to upload additional control or dual variables, while anchoring both perturbation and update directions to a more stable global direction.

\section{Training Process of FedAvg and FedSAM }\label{app:baselines}
As shown in \textbf{Algorithm~\ref{fedsam and fedavg}}, in each training round $t$, the PS first selects a subset, $\mathcal{S}^t \in \mathcal{N}$, of devices uniformly at random (UAR) for training. Considering limited bandwidth between the devices and PS, the subset of $S$ devices is selected, i.e., $|\mathcal{S}^t|=S$. Each training round consists of $K$ local training iterations at the $S$ participating devices. For local training iteration $k=0,1,\cdots,K-1$, each selected device $i \in \mathcal{S}^t$ performs a SAM-based local update, where gradient is evaluated at the perturbed local model $\tilde{\theta}_i^{t,k} = \theta_i^{t,k}+\epsilon_i^{t,k}$ instead of a vanilla SGD descent at the current local model ${\theta}_i^{t,k}$; see Steps 10--12. By generating gradients at the perturbed local model $\tilde{\theta}_i^{t,k}$, FedSAM smooths the local objective, and promotes solutions toward flatter minima that generalize better.
\begin{algorithm}[htbp]
    \caption{FedAvg \& FedSAM}
    \label{fedsam and fedavg}
    \begin{algorithmic}[1]
    \STATE {\textbf{Input:} $\theta^0$, $T$, $K$, $N$, $S$, $\rho$, $\eta$; }
    \FOR{$t=0,1,\cdots, T-1$}
    \STATE {The PS selects the device subset $\mathcal{S}^t \subseteq \mathcal{N}$ UAR, and sends $\theta^t$ to devices $i \in \mathcal{S}^t$; }
    \FOR{each device $i \in \mathcal{S}^t$ in parallel}
    \FOR{$k=0,1,\cdots, K-1$}
                \STATE {Compute $g_i^{t,k} = \nabla F_i(\theta_i^{t,k};\phi_i^{t,k}) $}
                \STATE {\textit{\textbf{\# FedAvg}}}
                \STATE {Update the local model with $\theta_i^{t,k+1} = \theta_i^{t,k} - \eta {g}_i^{t,k}$;}
                \STATE {\textbf{\textit{\# FedSAM}}}
                \STATE {Find $\epsilon_i^{t,k}$ via stochastic linearization $\epsilon_i^{t,k} \!=\! \rho \frac{g_i^{t,k}}{|| g_i^{t,k} ||}$;}
                \STATE { $\tilde{\theta}_i^{t,k} = \theta_i^{t,k} + \rho \frac{g_i^{t,k}}{ \left\|g_i^{t,k}\right\|}$, $\tilde{g}_i^{t,k} = \nabla F_i(\tilde{\theta}_i^{t,k}; \phi_i^{t,k})$;}
                \STATE {Update the local model with $\theta_i^{t,k+1} = \theta_i^{t,k} - \eta \tilde{g}_i^{t,k}$;}
            \ENDFOR
            \ENDFOR
            \STATE {The PS updates $\theta^t$ with $\theta^{t+1} = \frac{1}{S}\sum_{i \in \mathcal{S}^t} \theta_i^{t,K} $; }
            \ENDFOR
    \RETURN{${{\boldsymbol{\theta}}^{T}}$}
    \end{algorithmic}
\end{algorithm}
\section{Supplementary Experimental Results}
\subsection{Details on Algorithm Parameters} \label{app:algorithm-parameters}
Table~\ref{tab:algorithm-params} summarizes the parameter settings used in the experiments.
For a fair comparison, all methods use the same data partitions, device sampling, and random seeds under each dataset, heterogeneity level, and participation setting. Algorithm-specific hyperparameters are selected from the ranges recommended by the original papers.
For FedVSSAM, the mixing coefficients $(\gamma_l,\gamma_g)$ are chosen according to the heterogeneity level, and kept fixed across datasets and device sampling settings with the same heterogeneity level. 

\begin{table}[htbp]
\centering
\caption{Key Parameter Settings}
\label{tab:algorithm-params}
\small
\setlength{\tabcolsep}{3pt}
\renewcommand{\arraystretch}{1.08}
\begin{tabular}{@{}>{\centering\arraybackslash}p{0.13\textwidth}
>{\raggedright\arraybackslash}p{0.30\textwidth}
>{\raggedright\arraybackslash}p{0.49\textwidth}@{}}
\toprule
\textbf{Scope} & \textbf{Parameter} & \textbf{Setting} \\
\midrule
\multirow{4}{=}{\centering All methods} & Local benchmark optimizer & SGD \\
& Learning rate grids & $\eta_l$: $\{0.01,0.02,0.05,0.1,0.2\}$; \newline $\eta_g$: $\{0.5,1,2,3,4\}$ \\
& Batch size & CIFAR-10: 128; CIFAR-100: 20; DBpedia-14: 128 \\
& Local epochs & $K=10$ \\
\midrule
\multirow{2}{=}{\centering FedVSSAM} & $(\gamma_l,\gamma_g)$ & Moderate heterogeneity: $(0.4,0.6)$; \newline Severe heterogeneity: $(0.1,0.6)$ \\
& Perturbation radius & $\rho=0.05$ \\
\midrule
\multirow{2}{=}{\centering SAM-based FL baselines} & Perturbation-radius grid & $\rho:\{0.001,0.01,0.02,0.05,0.1,0.2\}$ \\
& Selected perturbation radius & $\rho=0.1$: FedSAM/MoFedSAM/FedSMOO; \newline $\rho=0.01$: FedLESAM; \newline $\rho=0.05$: FedGAMMA \\
\midrule
FedProx & Proximal coefficient & $\mu=0.1$ \\
MoFedSAM & Momentum coefficient & $\beta=0.1$ \\
FedCM & Client-level momentum & $\alpha=0.1$ \\
FedDyn & Dynamic regularization & $\mu=0.01$ \\
FedSMOO & Penalized coefficient & $\beta=10$ \\
FedACG & Local proximal coefficient & $\beta=0.01$ \\
FedAdam & Adam parameters & $(\beta_1,\beta_2)=(0.9,0.99)$ \\
FedAdam & Server parameters & $(\tau,\eta)$: CIFAR-10: $(0.001,0.01\!)$; CIFAR-100: $(0.1,1\!)$; DBpedia-14: $(0.001,0.01\!)$\\
\bottomrule
\end{tabular}
\end{table}

\subsection{Convergence Curves on CIFAR-10 and CIFAR-100}\label{app:convergence-curves}
\begin{figure}[H]
    \centering
	\subfloat[CIFAR-10, $\alpha=0.5$.]{\includegraphics[width=.24\textwidth]{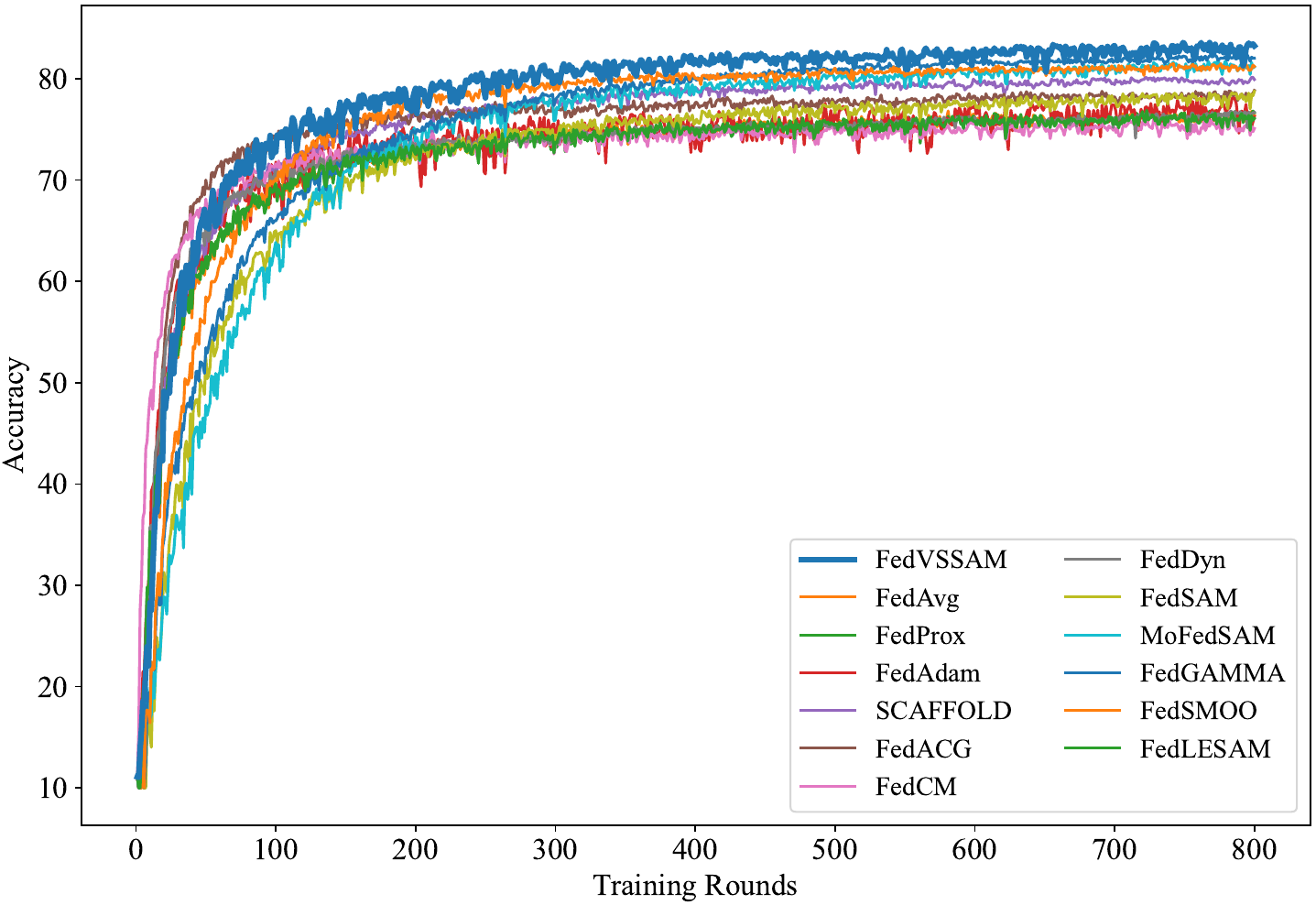}
		\label{fig_acc_1}} 
	\subfloat[CIFAR-10, $\alpha=0.1$.]{\includegraphics[width=.24\textwidth]{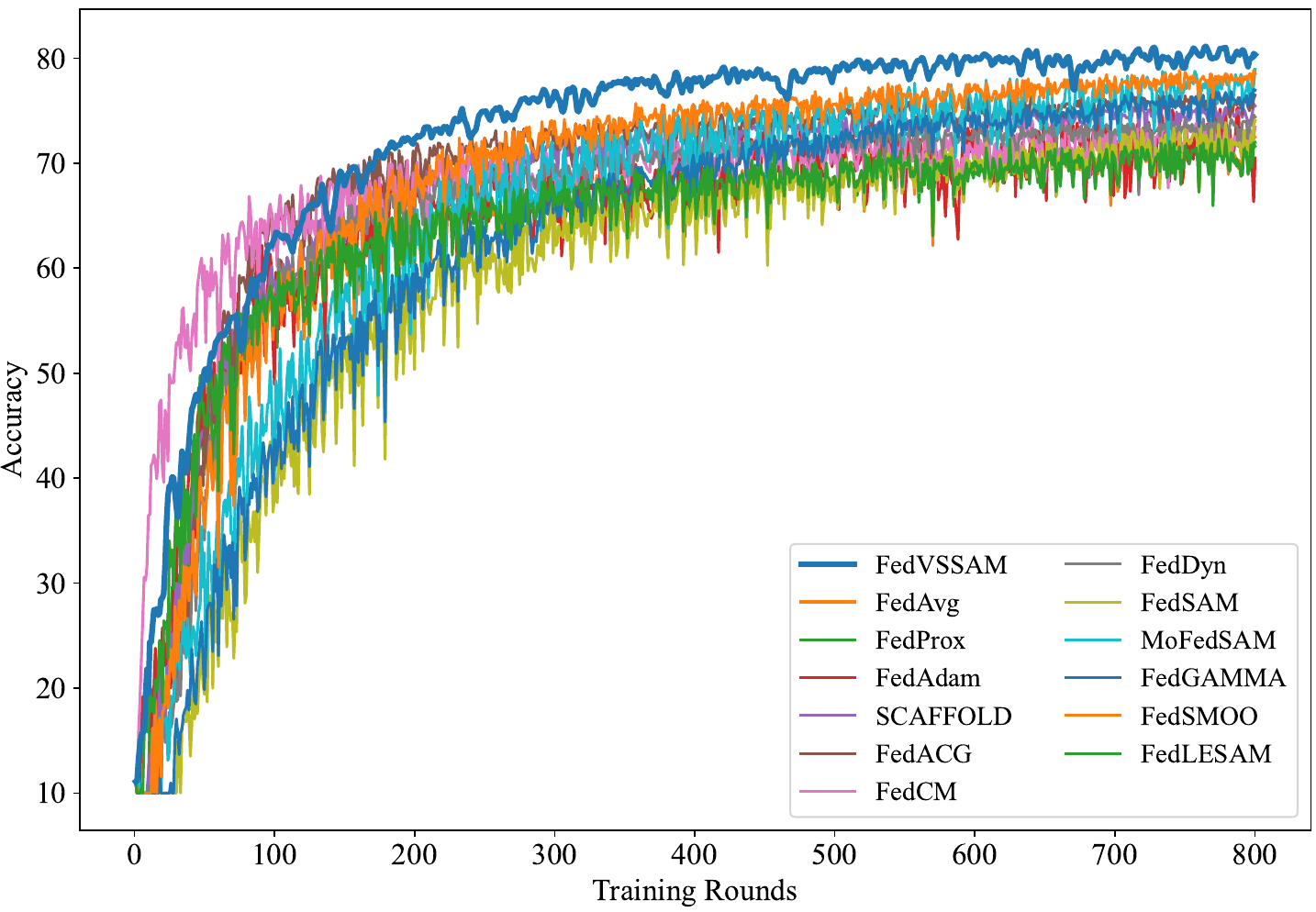}
		\label{fig_acc_2}}
    \subfloat[CIFAR-100, $\alpha=0.5$.]{\includegraphics[width=.24\textwidth]{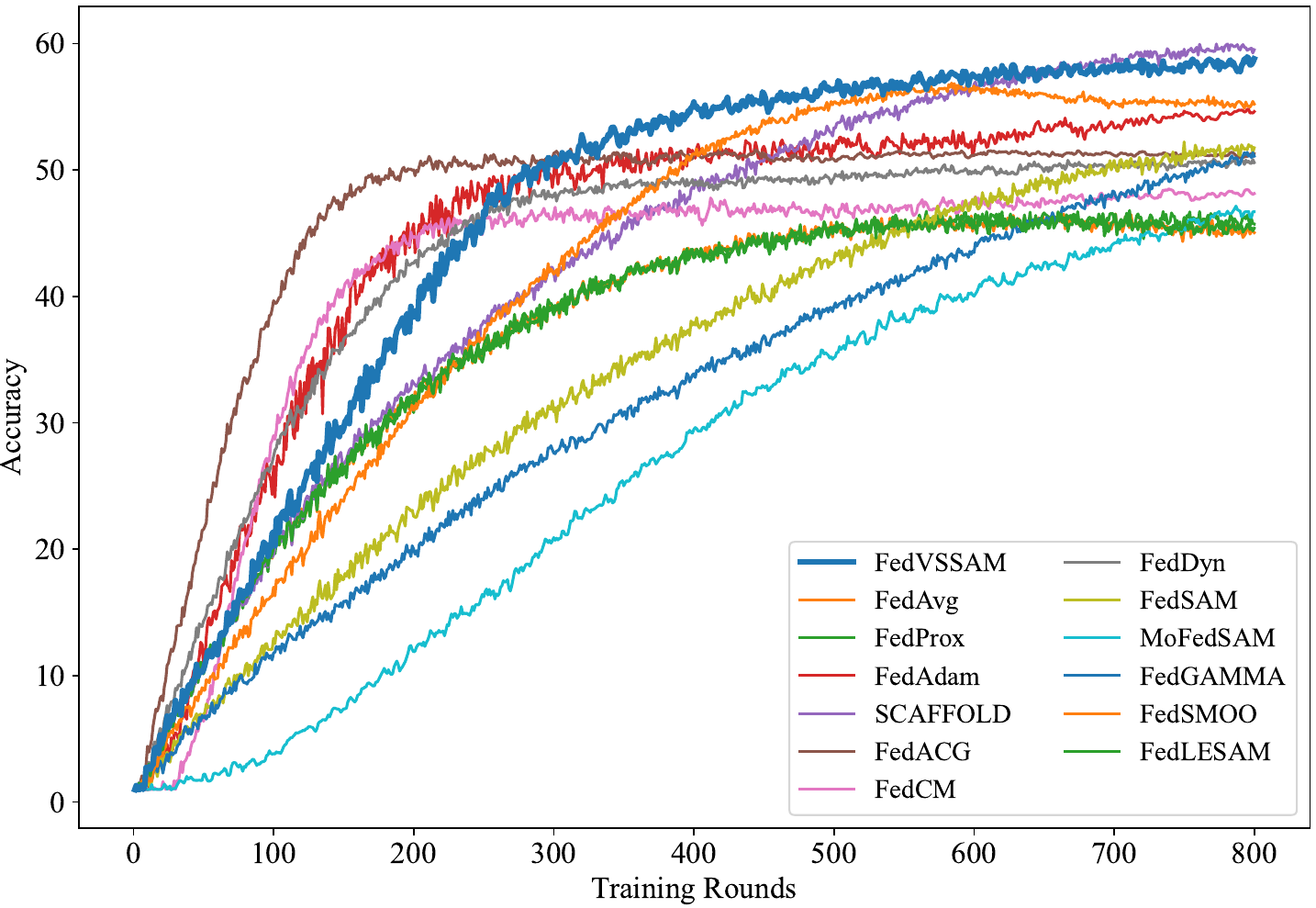}
    \label{fig_acc_3}}
    \subfloat[CIFAR-100, $\alpha=0.1$.]{\includegraphics[width=.24\textwidth]{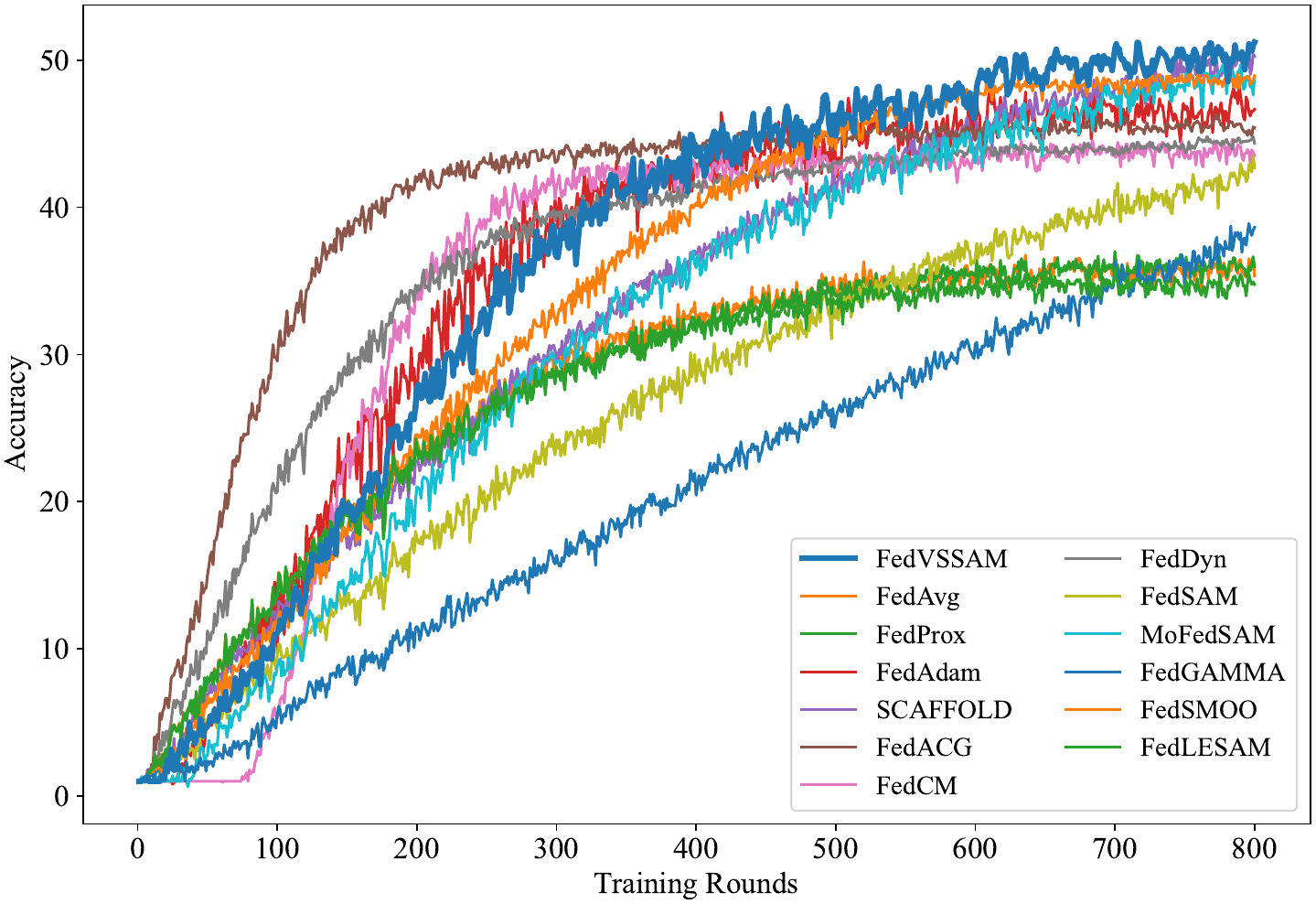}
    \label{fig_acc_4}}
	\caption{Convergence performance of different algorithms on CIFAR-10 and CIFAR-100.}
	\label{fig_acc}
\end{figure}

Fig.~\ref{fig_acc} reports the convergence curves on CIFAR-10 and CIFAR-100 under moderate and severe Dirichlet heterogeneity.
Compared with other SAM-based FL algorithms, the performance advantage of FedVSSAM comes from its direct treatment of flatness incompatibility. Instead of only introducing corrected or momentum-based local updates, FedVSSAM jointly suppresses the perturbation inconsistency and the local update variance through the variance-suppressed flatness search, local update, and global update, see \eqref{vs flatness search}, \eqref{vs local update}, and \eqref{vs global update}, respectively.
Thus, the local perturbations across devices are better aligned with a smoothed global direction, guiding the aggregated model toward a more globally consistent flat region under non-IID data and achieving better convergence and generalization.

\subsection{Additional Text Classification Results}\label{app:dbpedia-text}
To further evaluate FedVSSAM beyond image classification, we conduct additional text topic classification experiments on DBpedia-14 with a TextCNN model.
We use a balanced subset with 140,000 training samples and 14,000 test samples, corresponding to 10,000 training samples and 1,000 test samples per class.
The TextCNN uses a vocabulary size of 30,000, a maximum sequence length of 128, 100-dimensional embeddings, and 100 convolutional filters for each kernel size in $\{3,4,5\}$.
All methods are trained for 400 rounds with $K=10$, and the batch size is 128.

\begin{table}[H]
\centering
\caption{Test accuracy of FedVSSAM and baselines on the DBpedia-14 dataset after 400 training rounds. The device sampling setup is 10\%-100 devices. Acc. means the final accuracy, and Rd. represents the number of training rounds required for an algorithm to first reach a certain accuracy level, which is 90\%.
A dash (--) indicates that the corresponding target accuracy was not reached within the training budget under that setting.}
\label{tab:dbpedia-text}
\small
\setlength{\tabcolsep}{3pt}
\renewcommand{\arraystretch}{1.08}
\resizebox{0.5\textwidth}{!}{%
\begin{tabular}{l|cc|cc}
\toprule
\multirow{2}{*}{Algorithm} & \multicolumn{2}{c|}{Dirichlet, $\alpha=0.1$} & \multicolumn{2}{c}{Pathological, $c=3$} \\
\cline{2-5}
& Acc. & Rd. & Acc. & Rd. \\
\midrule
FedAvg~\cite{fedavg} 
& $92.56_{\pm .08}$ & 205 
& $91.09_{\pm .11}$ & 215 \\

FedProx~\cite{fedprox} 
& $92.40_{\pm .09}$ & 205 
& $91.01_{\pm .12}$ & 215 \\

FedAdam~\cite{reddi2021adaptive} 
& $96.19_{\pm .07}$ & \textbf{62} 
& $94.41_{\pm .09}$ & 56 \\

SCAFFOLD~\cite{scaffold} 
& $94.76_{\pm .10}$ & 133 
& $95.13_{\pm .08}$ & 113 \\

FedACG~\cite{kim2024communication} 
& $\mathbf{96.44}_{\pm .08}$ & \textbf{39} 
& $95.11_{\pm .11}$ & \textbf{46} \\

FedCM~\cite{xu2021fedcm} 
& $88.82_{\pm .16}$ & 320 
& $83.14_{\pm .21}$ & -- \\

FedDyn~\cite{feddyn} 
& $94.90_{\pm .12}$ & 96 
& $93.71_{\pm .10}$ & 93 \\

FedSAM~\cite{caldarola2022improving} 
& $93.93_{\pm .09}$ & 137 
& $93.54_{\pm .11}$ & 144 \\

MoFedSAM~\cite{qu2022generalized} 
& $94.43_{\pm .08}$ & 165 
& $94.46_{\pm .13}$ & 137 \\

FedGAMMA~\cite{10269141} 
& $95.76_{\pm .08}$ & 106 
& $95.94_{\pm .09}$ & 87 \\

FedSMOO~\cite{pmlr-v202-sun23h} 
& $96.16_{\pm .09}$ & 89 
& $\mathbf{96.21}_{\pm .11}$ & 80 \\

FedLESAM~\cite{pmlr-v235-fan24c} 
& $92.11_{\pm .10}$ & 205 
& $90.92_{\pm .12}$ & 215 \\

FedVSSAM(ours) 
& $\mathbf{96.90}_{\pm .07}$ & 71 
& $\mathbf{96.74}_{\pm .11}$ & \textbf{46} \\
\bottomrule
\end{tabular}%
}
\end{table}

\begin{figure}[H]
    \centering
	\subfloat[DBpedia-14, $\alpha=0.1$.]{\includegraphics[width=.3\textwidth]{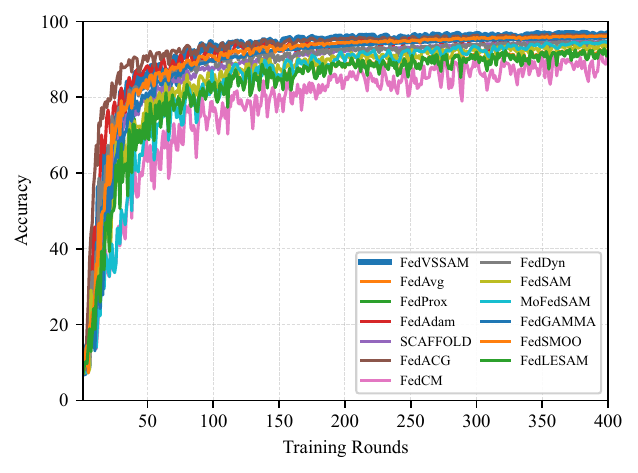}
		\label{fig_acc_1_db}} 
	\subfloat[DBpedia-14, $c=3$.]{\includegraphics[width=.3\textwidth]{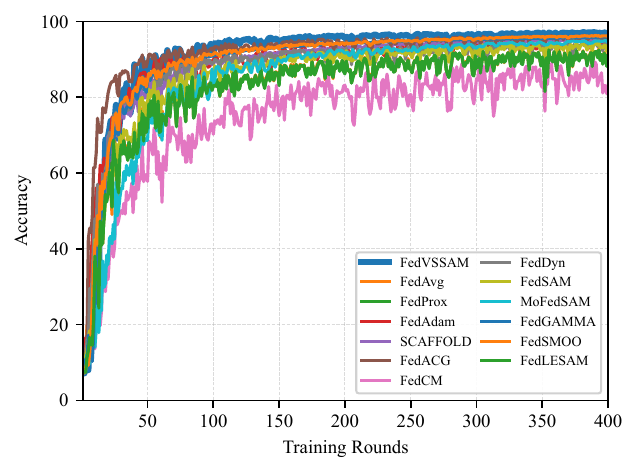}
		\label{fig_acc_2_db}}
	\caption{Convergence performance of different algorithms on DBpedia-14.}
	\label{fig_acc_db}
\end{figure}

As shown in Table~\ref{tab:dbpedia-text} and Fig.~\ref{fig_acc_db}, FedVSSAM achieves the best final test accuracy under both text-classification heterogeneity settings.
It is noted that FedACG accelerates early training by broadcasting a lookahead global gradient for directional momentum and reaches the 90\% threshold earlier than FedVSSAM on the Dirichlet partition. However, without an explicit mechanism to smooth the sharp loss landscapes typical of skewed text data, FedACG fails to manage incompatible local minima across devices, leading to a lower final performance than FedVSSAM.
Under the Pathological partition setting, FedVSSAM obtains 96.74\% accuracy, while also reaching 90\% accuracy in 46 rounds, matching the fastest baseline FedACG.
These results are consistent with the role of variance-suppressed flatness search and updates: under topic-skewed text data, different devices observe substantially different word and phrase distributions, and FedVSSAM better aligns the perturbation and update directions across devices.

\subsection{t-SNE Visualization of Learned Representations}
\label{app:tsne-visualization}
To further inspect the representations learned by different algorithms, we visualize the feature embeddings of the global models using t-SNE. We compare FedAvg, SCAFFOLD, FedSAM, FedSMOO, and FedVSSAM on CIFAR-10 with ResNet-18 and Dirichlet distribution with $\alpha=0.1$, and use class labels only for coloring the embedded test samples with a size of 3,000.

\begin{figure}[htbp]
    \centering
    \subfloat[After 400 training rounds.]{
        \includegraphics[width=0.95\textwidth]{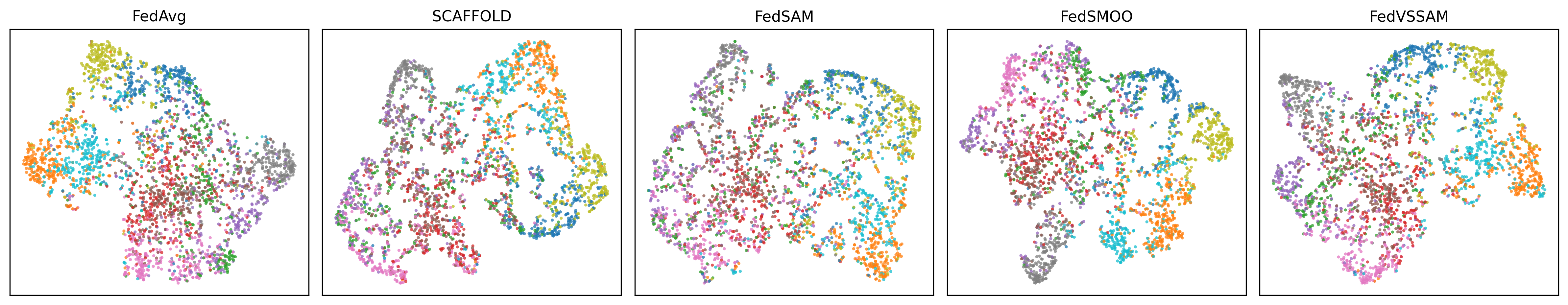}
        \label{fig:tsne-cifar10-400}
    }\\[1mm]
    \subfloat[After 800 training rounds.]{
        \includegraphics[width=0.95\textwidth]{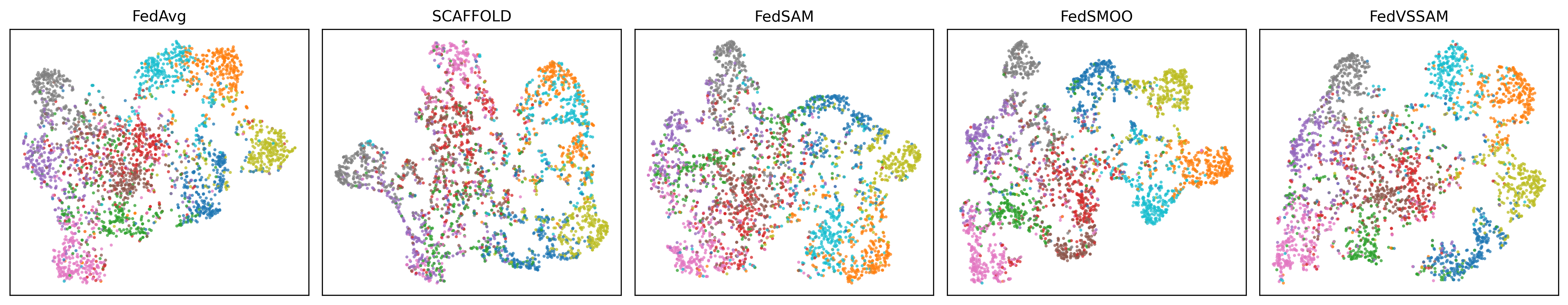}
        \label{fig:tsne-cifar10-800}
    }
    \caption{t-SNE visualization of global model feature embeddings on CIFAR-10 with ResNet-18. Each color denotes one class.}
    \label{fig:tsne-cifar10}
\end{figure}

Here, the visualization is intended as a supplementary measure of generalization performance.
Fig.~\ref{fig:tsne-cifar10} shows the t-SNE embeddings after 400 and 800 training rounds.
At 400 rounds, FedAvg, SCAFFOLD, and FedSAM still exhibit substantial class mixing in the embedding space, indicating that the learned features are not yet clearly discriminative under heterogeneous FL.
FedSMOO and FedVSSAM show more visible class-wise structures, while FedVSSAM already forms several more coherent peripheral clusters and reduces part of the central class overlap.

After 800 rounds, the separation of class clusters becomes clearer for all methods, but the difference among methods remains visible.
FedVSSAM produces relatively more compact and better separated class groups than FedAvg, SCAFFOLD, and FedSAM, suggesting that the proposed variance-suppressed flatness search, local update, and global update help learn more discriminative global representations.
FedSMOO also yields clear cluster structures in several regions, which is consistent with its robust empirical performance.
However, FedVSSAM achieves comparable representation separation while retaining the communication advantage discussed in Appendix~\ref{app:conv-comm-comparison}.
These visualizations are consistent with the main experimental results: stabilizing the directions used for local flatness search, local update, and global update can improve the quality of the learned global representation under data heterogeneity.

\subsection{Results on Global Direction Tracking and Flatness Incompatibility Suppression}
\label{app:flatness-incompatibility-probe}

To further verify the mechanism of FedVSSAM beyond test accuracy, we conduct an additional experiment on CIFAR-10 with ResNet-18 under Dirichlet distribution with $\alpha=0.1$ and $\alpha=0.5$. The system contains $100$ devices with $10\%$ devices sampled UAR per round. The mechanism metrics are evaluated every 80 training rounds. We use a fixed subset of 50 devices to evaluate the local flatness, and construct an IID reference gradient from 10 randomly sampled batches of data with a batch size of 512 from the CIFAR-10 training distribution.

\begin{figure}[htbp]
	\centering
	\begin{minipage}{0.49\textwidth}
		\centering
		\subfloat[$\alpha=0.5$.]{
			\includegraphics[width=.48\linewidth]{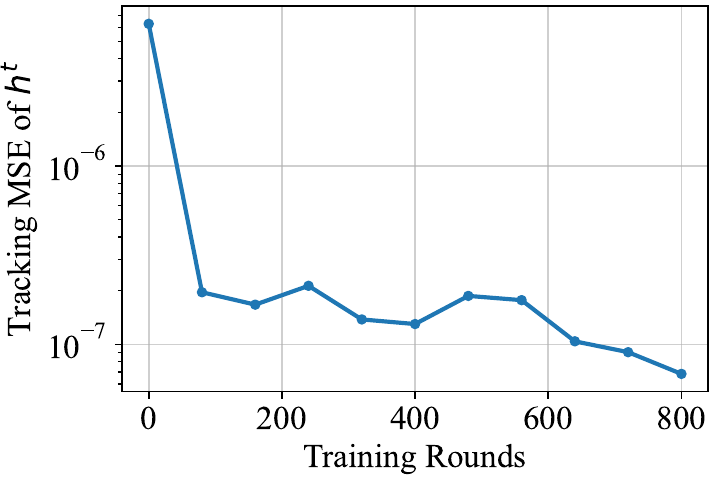}
			\label{fig:mechanism-direction-tracking-05}}
		\subfloat[$\alpha=0.1$.]{
			\includegraphics[width=.48\linewidth]{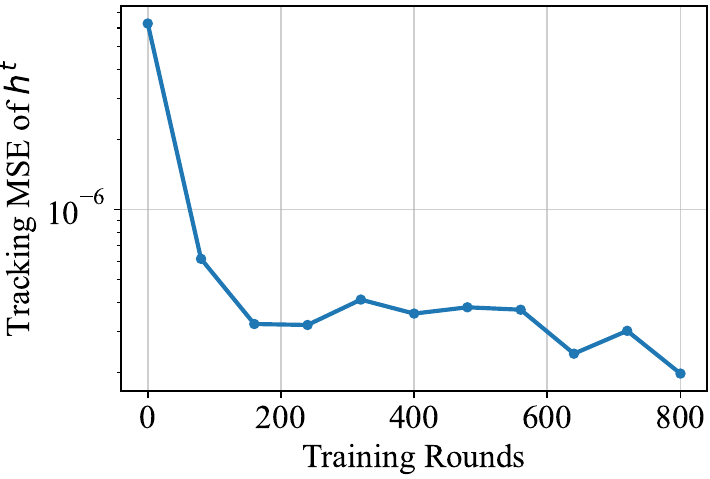}
			\label{fig:mechanism-direction-tracking-01}}
		\caption{Tracking error between $h^t$ and the global gradient for FedVSSAM.}
		\label{fig:mechanism-direction-tracking}
	\end{minipage}\hfill
	\begin{minipage}{0.49\textwidth}
		\centering
		\subfloat[$\alpha=0.5$.]{
			\includegraphics[width=.48\linewidth]{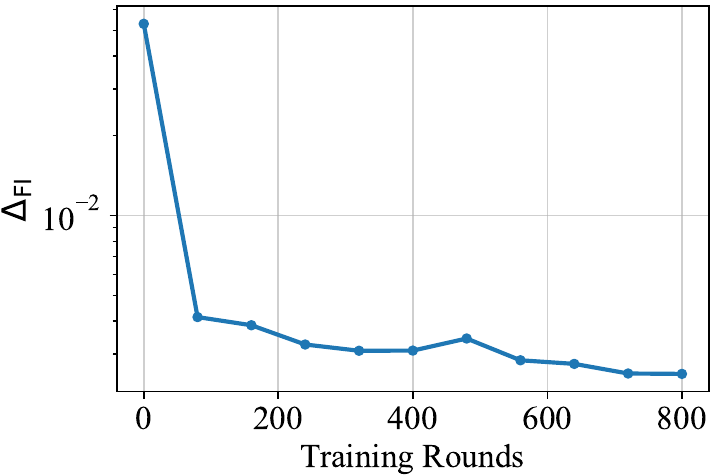}
			\label{fig:mechanism-flatness-proxy-05}}
		\subfloat[$\alpha=0.1$.]{
			\includegraphics[width=.48\linewidth]{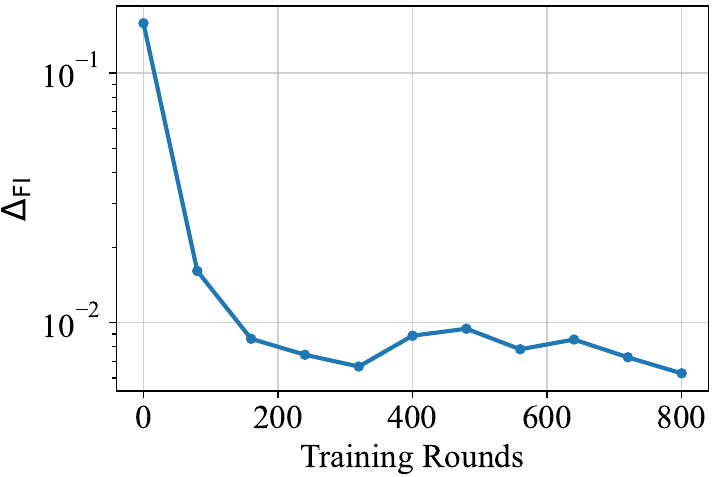}
			\label{fig:mechanism-flatness-proxy-01}}
		\caption{The flatness incompatibility proxy value of FedVSSAM.}
		\label{fig:mechanism-flatness-proxy}
	\end{minipage}
\end{figure}

First, let $\widehat{\nabla F}_{\mathrm{iid}}(\theta^t)$ denotes the empirical gradient estimated from IID sampled batches of data at the current global model $\theta^t$. For FedVSSAM, the corresponding global direction is the variance-suppressed direction $h^t$. 
Fig.~\ref{fig:mechanism-direction-tracking} shows that FedVSSAM achieves the tracking error $\|h^t-\widehat{\nabla F}_{\mathrm{iid}}(\theta^t)\|^2$ keeps decreasing as the training progresses, which is consistent with the variance-suppression property of $h^t$ established in \textbf{Theorem~2}.

We also illustrate FedVSSAM's ability to suppress flatness incompatibility. Let $\mathcal{N}_{\mathrm{eval}}\subseteq\mathcal{N}$ denote the fixed subset of evaluation devices. 
For each evaluation device $i\in\mathcal{N}_{\mathrm{eval}}$, we first estimate the local empirical gradient $\hat g_i^t$ at $\theta^t$. 
Following the FedVSSAM search direction in \eqref{vs flatness search}, we define $d_i^t=(1-\gamma_l)h^t+\gamma_l \hat g_i^t$, which is used to estimate the local flatness proxy $s_i^t(\theta^t,\rho)=F_i\!\left(\theta^t+\rho\frac{d_i^t}{\|d_i^t\|}\right)-F_i(\theta^t)$.
We then compute the empirical flatness incompatibility as
\begin{align}
\widehat{\Delta}_{\mathrm{FI}}^t(\theta^t,\rho)=\frac{1}{|\mathcal{N}_{\mathrm{eval}}|}\sum_{i\in\mathcal{N}_{\mathrm{eval}}}\Big(s_i^t(\theta^t,\rho)-\frac{1}{|\mathcal{N}_{\mathrm{eval}}|}\sum_{j\in\mathcal{N}_{\mathrm{eval}}}s_j^t(\theta^t,\rho)\Big)^2 ,
\end{align}
where a smaller value of $\widehat{\Delta}_{\mathrm{FI}}^t(\theta^t,\rho)$ indicates that different devices perceive more consistent local flat regions around the current global model.

Fig.~\ref{fig:mechanism-flatness-proxy} reports the flatness incompatibility proxy. It is noted that FedVSSAM yields a low flatness incompatibility proxy as training progresses, supporting that the shared variance-suppressed direction reduces the incompatibility of local flatness estimates under different levels of data heterogeneity.
The above results provide direct empirical evidence for the proposed mechanism: FedVSSAM suppresses the dispersion of local flatness proxies and aligns local SAM search directions with a more stable global direction, and the performance gain is accompanied by a reduction in the flatness incompatibility defined in Section~\ref{subsec:flat-incompat}.

\subsection{Sensitivity Analysis}\label{app:sensitivity}
\begin{figure}[H]
	\centering
	\subfloat[$\gamma_l$.]{\includegraphics[width=.24\textwidth]{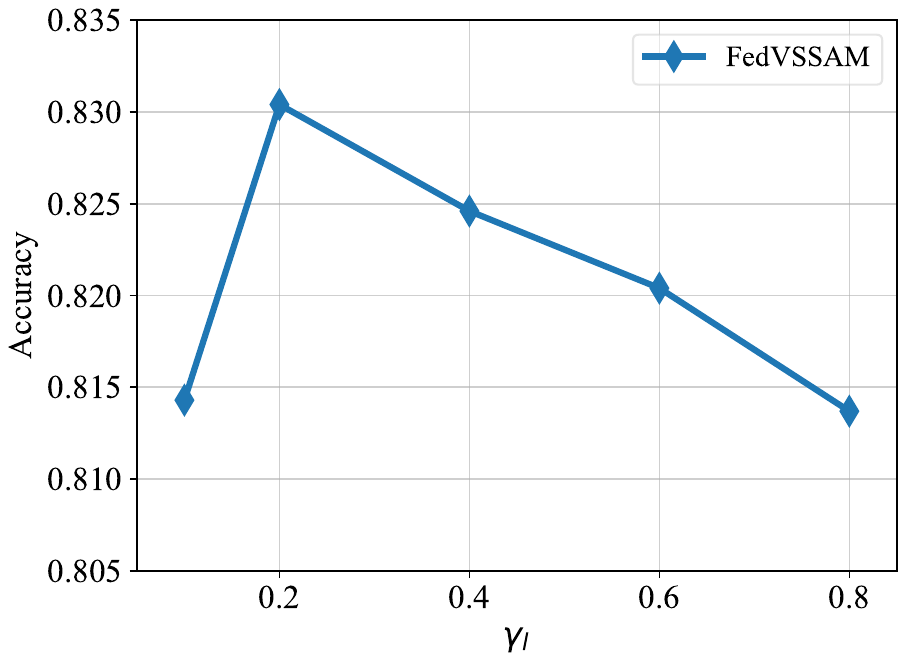}
		\label{gamma_l}} 
	\subfloat[$\gamma_g$.]{\includegraphics[width=.24\textwidth]{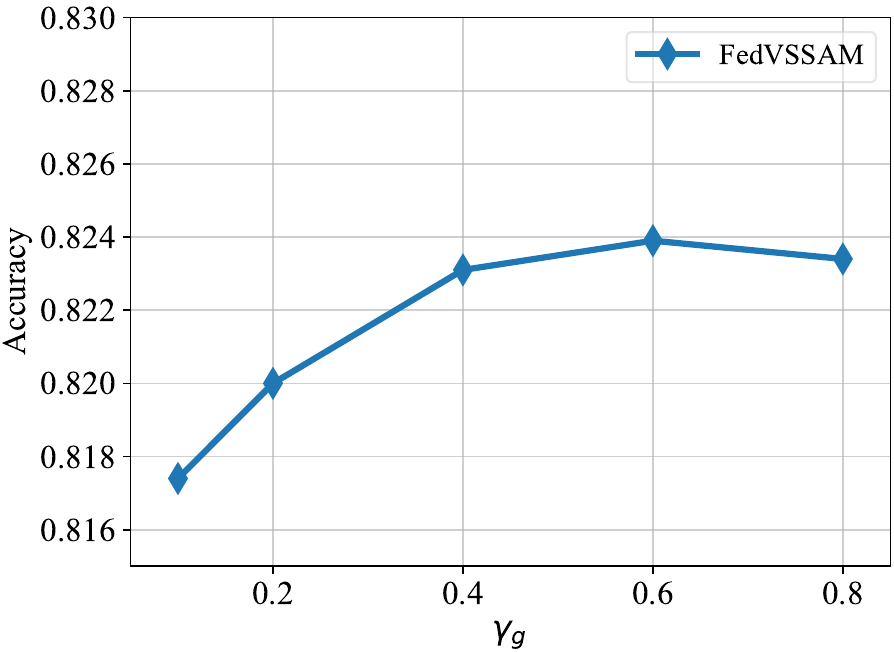}
		\label{gamma_g}}
    \subfloat[Local epochs $K$.]{\includegraphics[width=.24\textwidth]{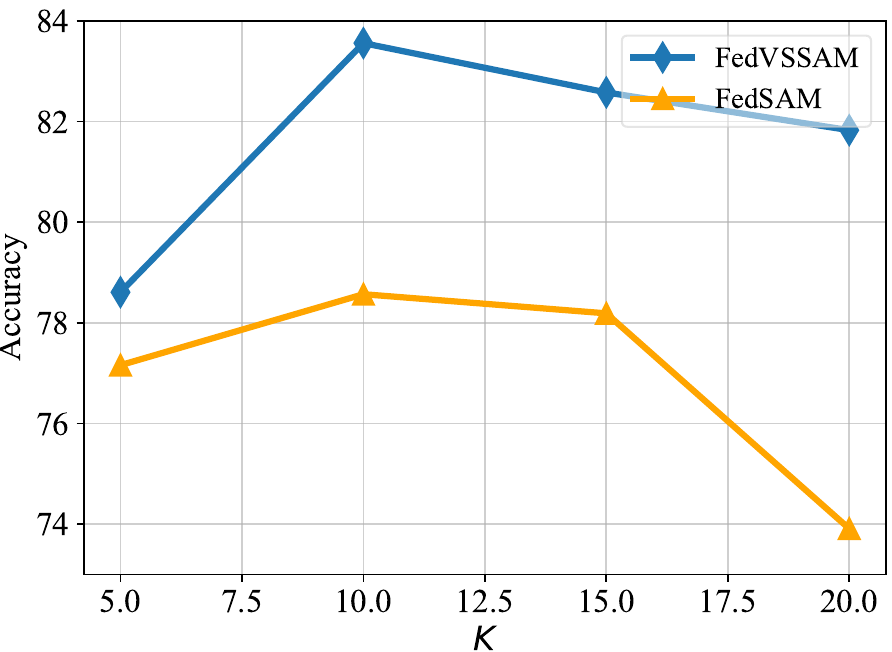}
    \label{K}}
    \subfloat[Radius $\rho$.]{\includegraphics[width=.24\textwidth]{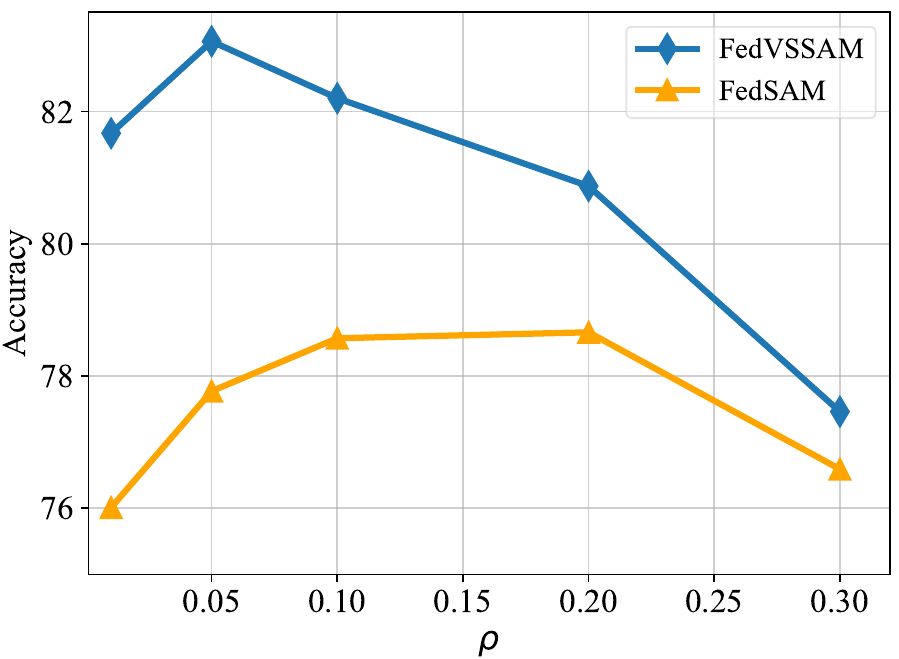}
    \label{rho}}
	\caption{Sensitivity of FedVSSAM to different parameters on the CIFAR-10 dataset with $\alpha=0.5$.}
	\label{fig_ablation}
\end{figure}

We examine the sensitivity of FedVSSAM to several key parameters: the mixing coefficients $\gamma_l$ and $\gamma_g$, the number $K$ of local training epochs, and the perturbation radius $\rho$.
As shown in Fig.~\ref{fig_ablation}, the robustness of FedVSSAM is assessed through variations in the hyperparameters by adjusting each parameter while keeping the others unchanged.
For the mixing coefficients, taking $\gamma_l \rightarrow 0$ can be overly conservative: since $h^t$ is a smoothed and history-dependent global direction, and can be biased under partial participation, relying almost exclusively on $h^t$ may render the perturbation stale with respect to the current local iterate $\theta_i^{t,k}$. As a consequence, the local perturbation may no longer approximate the steepest local ascent direction required by SAM's inner maximization, weakening curvature adaptivity and slowing down progress. In contrast, an excessively large $\gamma_l$ makes the local flatness search and update dependent excessively on instantaneous local stochastic gradients, hence increasing the impact of data heterogeneity and friendly adversary. Consequently, a moderate $\gamma_l$ provides the balanced trade-off between local adaptivity and variance suppression.
A similar trade-off is observed for $\gamma_g$. Particularly, an excessively small $\gamma_g$ causes the PS update of $h^t$ to be excessively slow, whereas an excessively large $\gamma_g$ weakens the smoothing effect on aggregation noise. This is consistent with the role of $\gamma_l\gamma_g$ revealed in \textbf{Theorem 2}, which balances the vanishing term and residual variance in \eqref{the good of variance suppression eq}. Increasing the number of local training epochs $K$ initially improves the training of FedVSSAM by allowing more effective local progress within each training round, but a large $K$ enlarges local drift.
Moreover, a moderate perturbation radius $\rho$ is critical: A small $\rho$ weakens the flatness-seeking effect, while an excessively large $\rho$ amplifies the mismatch among perturbed local models, which is consistent with \textbf{Lemma~1}.

\section{Comparison of Convergence Rates and Communication Requirements}\label{app:conv-comm-comparison}
In this paper, we use $T$ for training rounds, $K$ for local training iterations, $S$ for selected devices, $N$ for all devices, and $\Delta= F(\tilde{\theta}^0)-F^*$ for the initial optimality gap. 
Since existing convergence guarantees are established under different technical assumptions, the comparison is not solely intended as a ranking of rates. 
Instead, it separates two aspects: the communication or auxiliary-state cost required to implement each correction mechanism, and the residual error terms that remain explicit in its convergence guarantee.

Table~\ref{tab:conv-rate-comparison} compares the reported stationarity bounds together with the auxiliary uplink vector required by each algorithm.
SCAFFOLD, FedGAMMA, FedLESAM-S, and FedSMOO use control variates or dynamic regularization, so their bounds keep correction-state or correction-inconsistency terms, and the corresponding implementations require an additional uplink vector.
MoFedSAM, FedGF, FedGMT, and FedWMSAM avoid this uplink cost, but their bounds still isolate terms associated with momentum weighting, global perturbation estimation, global trajectory or ADMM inconsistency, or perturbation-induced variance.
FedVSSAM keeps the device upload identical to a local model update.
At round t, the PS broadcasts $h^t$ to selected devices; after aggregation, it updates $h^{t+1}$ for the next round.
The same adjusted direction is then used in local flatness search, local update, and global update, so the bound of FedVSSAM does not introduce an isolated correction-specific residual beyond the common stochastic gradient, local update, heterogeneity, and partial participation terms.

\textbf{Corollary}~\ref{convergence corollary} shows how the common error sources appear in the bound of FedVSSAM.
The term $\sqrt{L\Delta\sigma_{l,1}^2/(NKT)}$ reflects stochastic gradient deviation in local training, and decreases with a larger scale of system devices.
The term $\sqrt{L\Delta\sigma^2/(ST)}$, where $\sigma^2=\sigma_{l,1}^2+\sigma_{g,1}^2$, reflects the UAR sampling variance, including both local stochastic gradient variance and cross-device heterogeneity.
\textbf{Theorem}~\ref{the good of variance suppression} further complements this result by bounding the tracking error of $h^t$.
It controls the MSE between the adjusted direction and the global gradient, which is consistent with the design in \eqref{vs flatness search}--\eqref{vs global update}: $h^t$ serves as a common anchor for local flatness search, perturbed local update, and the global update.

\begin{table*}[t]
\centering
\caption{Comparison of reported non-convex convergence guarantees and auxiliary uplink vector requirements. Rates are normalized to the notation of this paper when possible, with method-specific constants and lower-order terms suppressed. N/A indicates that a comparable stationarity bound is not reported in the cited work. A dash (--) means that the bound does not induce an additional residual error beyond the common stochastic gradient, local update, heterogeneity, or partial participation errors considered in this paper; it does not mean that the method has no residual error.}

\label{tab:conv-rate-comparison}
\begingroup
\fontsize{6.6pt}{7.45pt}\selectfont
\setlength{\tabcolsep}{2.0pt}
\renewcommand{\arraystretch}{1.12}
\resizebox{0.985\textwidth}{!}{%
\begin{tabular}{@{}
>{\raggedright\arraybackslash}p{0.120\textwidth}
l 
>{\raggedright\arraybackslash}p{0.255\textwidth}
>{\centering\arraybackslash}p{0.060\textwidth}
@{}}
\toprule
\textbf{Method}
& \textbf{\shortstack{Reported\\stationarity\\bound}}
& \textbf{\shortstack{Specific\\residual\\error}}
& \textbf{\shortstack{Extra\\uplink\\vector}} \\
\midrule

FedAvg~\cite{fedavg}
& $\mathcal{O}\!\left(1/\sqrt{KT}\right)+\mathrm{device drift\;term}$
& --
& No \\

SCAFFOLD~\cite{scaffold}
& $\tilde{\mathcal{O}}\!\left(\sqrt{\frac{L\Delta\sigma^2}{SKT}}+\frac{L\Delta}{T}\left(\frac{N}{S}\right)^{2/3}\right)$
& Control variate tracking error
& Yes \\

MoFedSAM~\cite{qu2022generalized}
& $\mathcal{O}\!\left(\frac{\beta L\Delta}{\sqrt{TKS}}+\frac{\beta\sqrt{K}\sigma_g^2}{\sqrt{TS}}+\frac{L^2\sigma_l^2}{T^{3/2}K}+\frac{\sqrt{K}L^2}{T^{3/2}\sqrt{S}}\right)$
& --; momentum-weighted common variance terms
& No \\

FedGAMMA~\cite{10269141}
& $\mathcal{O}\!\left(1/\sqrt{TKS}\right)+\mathcal{E}_{\mathrm{GAMMA}}$
& Control variate tracking error
& Yes \\

FedLESAM-S~\cite{pmlr-v235-fan24c}
& $\mathcal{O}\!\left(
\frac{L\Delta}{\sqrt{TKN}}
+\frac{L^2\rho^2\sigma_g^2}{TK}
+\frac{L^2\rho^2}{T}
+\frac{L^2\sigma_l^2\rho^2}{\sqrt{TKN}}
\right)+\mathcal{E}_{\mathrm{Scaf}}$
& Global perturbation estimation and control variate tracking error
& Yes \\

FedSMOO~\cite{pmlr-v202-sun23h}
& $\frac{1}{\zeta\beta T}\!\left(
\kappa_f
+\frac{3\beta L\kappa_r N}{S}
+\frac{72\beta^2L^2\delta^0 N}{S}
\right)$
& Initial correction inconsistency error
& Yes \\

FedGF~\cite{pmlr-v235-lee24aa}
& $\begin{gathered}
\mathcal{O}\!\left(
\frac{L\Delta}{\sqrt{TKS}}
+\left(\frac{(1-c)^2}{T}+1\right)\sigma_g^2
+\frac{L^2(1-c)^2\sigma_l^2}{T^{3/2}\sqrt{KS}}
\right.\\[-0.6mm]
\left.
+\frac{L^2(c^2\epsilon^2+1)}{T}
\right)
\end{gathered}$
& Global perturbation estimation error
& No \\

FedGloSS~\cite{Caldarola_2025_CVPR}
& N/A
& PS-side global-sharpness approximation; no comparable stationarity residual is reported
& No \\

FedGMT~\cite{pmlr-v267-li25bd}
& $\frac{1}{\kappa T}\!\left(
L(w^1)-L^*
+\frac{20M\beta^2L^2}{N}\Phi^0
+\frac{18\beta G}{N}
\right)$
& Global trajectory and ADMM inconsistency error
& No \\

FedWMSAM~\cite{li2025fedwmsam}
& $\tilde{\mathcal{O}}\!\left(\sqrt{\frac{L\Delta\sigma_\rho^2}{SKT}}+\frac{L\Delta}{T}\left(1+\frac{N^{2/3}}{S}\right)\right)$
& Perturbation-induced variance
& No \\

\textbf{FedVSSAM}
& $\mathcal{O}\!\left(
\frac{L\Delta}{T}
+\sqrt{\frac{L\Delta\sigma_{l,1}^2}{NKT}}
+\sqrt{\frac{L\Delta\sigma^2}{ST}}
\right)$
& --
& \textbf{No} \\
\bottomrule
\end{tabular}
}
\endgroup
\end{table*}

\begin{table*}[t]
\centering
\caption{Comparison of design properties for mitigating flatness incompatibility in SAM-based FL. Each entry indicates whether the corresponding property is explicitly addressed in the algorithms.}
\label{tab:communication-comparison}
\footnotesize
\setlength{\tabcolsep}{4pt}
\renewcommand{\arraystretch}{1.12}
\resizebox{\textwidth}{!}{%
\begin{tabular}{@{}>{\raggedright\arraybackslash}p{0.18\textwidth}
>{\centering\arraybackslash}p{0.14\textwidth}
>{\centering\arraybackslash}p{0.13\textwidth}
>{\centering\arraybackslash}p{0.16\textwidth}
>{\centering\arraybackslash}p{0.17\textwidth}
>{\centering\arraybackslash}p{0.17\textwidth}@{}}
\toprule
\textbf{Method} & \textbf{\shortstack{Non-IID\\Correction}} & \textbf{\shortstack{Update Bias\\Control}} & \textbf{\shortstack{SAM Perturbation\\Noise Aware}} & \textbf{\shortstack{Global Flatness\\Aligned}} & \textbf{\shortstack{Stage Coupled\\Direction}} \\
\midrule
FedAvg~\cite{fedavg} 
& No & No & No & No & No \\
SCAFFOLD~\cite{scaffold} 
& Yes & Yes & No & No & No \\
FedSAM~\cite{caldarola2022improving} 
& No & No & No & No & No \\
MoFedSAM~\cite{qu2022generalized} 
& Yes & Yes & No & No & No \\
FedGAMMA~\cite{10269141} 
& Yes & Yes & Yes & Yes & No \\
FedLESAM-S~\cite{pmlr-v235-fan24c} 
& Yes & Yes & Yes & Yes & No \\
FedSMOO~\cite{pmlr-v202-sun23h} 
& Yes & Yes & Yes & Yes & No \\
FedGF~\cite{pmlr-v235-lee24aa} 
& Yes & No & Yes & Yes & No \\
FedGloSS~\cite{Caldarola_2025_CVPR} 
& Yes & Yes & Yes & Yes & No \\
FedGMT~\cite{pmlr-v267-li25bd} 
& Yes & Yes & Yes & Yes & No \\
FedWMSAM~\cite{li2025fedwmsam} 
& Yes & Yes & Yes & Yes & No \\
\textbf{FedVSSAM} 
& \textbf{Yes} & \textbf{Yes} & \textbf{Yes} & \textbf{Yes} & \textbf{Yes} \\
\bottomrule
\end{tabular}
}
\end{table*}

In Table~\ref{tab:communication-comparison}, ``Non-IID Correction'' means that the method explicitly introduces a mechanism beyond vanilla local training to mitigate data heterogeneity or device drifts. 
``Update Bias Control'' indicates whether the method explicitly regularizes or corrects the bias of local update directions with respect to the global objective, beyond only modifying the sharpness objective.
``SAM Perturbation Noise Aware'' refers to whether the method explicitly handles the stochasticity or approximation error in the sharpness-aware perturbation or sharpness-related direction, rather than SGD noise only.
``Global Flatness Aligned'' means that the perturbation or sharpness objective is tied to the global loss landscape, not only to a device's local loss. 
``Stage Coupled Direction'' indicates whether the same stabilized direction is used across the perturbation search, the local update, and the global update. 

It is noted that FedGAMMA, FedGF, FedLESAM-S, FedGloSS, FedSMOO, FedGMT, and FedWMSAM each address part of the global-flatness problem. 
Specifically, FedGAMMA introduces global sharpness-aware correction, FedGF and FedLESAM-S improve the perturbation estimate toward the global loss landscape, FedGloSS moves sharpness optimization to the server side and approximates the global perturbation using the previous pseudo-gradient, FedSMOO relies on dynamic regularization, FedGMT measures sharpness along the global model trajectory, and FedWMSAM uses momentum-based perturbation with an adaptive momentum--SAM schedule. 
These designs can reduce the incompatibility between 
local and global flatness, but they do not reuse one common stabilized direction across perturbation construction, perturbed local update, and global update. 
By contrast, FedVSSAM attaches the correction to the adjusted direction $h^t$, which is used to construct $h_i^{t,k}$ in \eqref{vs flatness search}, to form $u_i^{t,k}$ in \eqref{vs local update}, and to update $h^{t+1}$ through the EMA recursion in \eqref{vs global update}.
Therefore, the ``Stage Coupled Direction'' entry in Table~\ref{tab:communication-comparison} and the MSE bound on $\|h^t-\nabla F(\theta^t)\|^2$ characterize the same mechanism: FedVSSAM controls the error of the direction that is reused throughout the SAM-based FL update, rather than introducing an isolated correction for only one stage.

\section{Proof of Lemma 1}\label{app:proof-lemma1}
With Taylor expansion of $s_i(\theta,\rho), \forall i \in \mathcal{N}$, we first have 
\begin{align} \label{s_i bound}
s_i(\theta,\rho) &= \rho\,\big\langle \nabla F_i(\theta), u_i \big\rangle + r_i \nonumber \\
&= \rho \|\nabla F_i(\theta)\| + \rho\langle \nabla F_i(\theta), u_i-d_i \rangle + r_i,
\end{align}
where $u_i = \frac{g_i}{\left\|g_i\right\|}$, $d_i = \frac{\nabla F_i\left(\theta\right)}{\left\|\nabla F_i\left(\theta\right)\right\|} $, and $r_i$ denotes the remainder with $\big\|r_i\big\|\le \tfrac{L}{2}\rho^2$.
By substituting \eqref{s_i bound} into \eqref{flatness incompatibility definition}, we obtain
\begin{align} \label{eq:A-step2}
\Delta_{\mathrm{FI}}(\theta,\rho) 
&\overset{(a)}{\leq} \frac{3\rho^2}{N}\!\sum_{i \in \mathcal{N}} \left[
\big(\|\nabla F_i(\theta)\| - m \big)^2 \right. \nonumber \\
&+ \left. \|\nabla F_i(\theta)\|^2 \|u_i-d_i\|^2
\!+\! \big(\frac{r_i-\bar{r}}{\rho}\big)^2 \right], \
\end{align}
where $m = \frac{1}{N}\sum_{j \in \mathcal{N}} \| \nabla F_j(\theta) \|$ and $\bar{r}=\frac{1}{N}\sum_{j \in \mathcal{N}} r_j$. Moreover, $(a)$ follows from the Cauchy-Schwarz inequality.

By choosing $m'=\|\nabla F(\theta)\|$, it follows that
\begin{align}
&\frac{1}{N}\sum_{i \in \mathcal{N}}\Big(\|\nabla F_i(\theta)\|-m\Big)^2 \overset{(a)}{\leq} \frac{1}{N}\sum_{i \in \mathcal{N}}\Big(\|\nabla F_i(\theta)\|-m'\Big)^2 \nonumber \\
& \overset{(b)}{\leq} \frac{1}{N}\sum_{i \in \mathcal{N}} \big\|\nabla F_i(\theta)-\nabla F(\theta)\big\|^2 = \frac{1}{N}\sum_{i \in \mathcal{N}} \|\delta_i(\theta)\|^2 ,
\end{align}
where $(a)$ holds since $\frac{1}{N}\sum_{i \in \mathcal{N}} ( \|\nabla F_i(\theta)\| -m' )^2
= \frac{1}{N}\sum_{i \in \mathcal{N}} (\|\nabla F_i(\theta)\|-m)^2 + (m-m')^2$, and $(b)$ is due to $|\|\boldsymbol{x}\|-\|\boldsymbol{y}\||\le \|\boldsymbol{x}-\boldsymbol{y}\|$.

For the term $\|u_i-d_i\|$ in \eqref{eq:A-step2}, we have
\begin{align}
   & \|u_i \!-\! d_i\| \!=\! \left\| \frac{g_i}{\left\|g_i\right\|} \!-\! \frac{\nabla F_i\left(\theta\right)}{\left\|\nabla F_i\left(\theta\right)\right\|} \right\| \overset{(a)}{\leq} \frac{2\| g_i \!-\! \nabla F_i\left(\theta\right)  \| }{\|\nabla F_i\left(\theta\right)\|} ,
\end{align}
where $(a)$ comes from $\left\| \|\boldsymbol{y}\| \boldsymbol{x}-\|\boldsymbol{x}\| \boldsymbol{y} \right\| \leq 2\|\boldsymbol{y}\|\|\boldsymbol{x}-\boldsymbol{y}\| $.

Since $\big\|r_i\big\|\le \tfrac{L}{2}\rho^2$, we have $\frac{1}{N}\sum_{i \in \mathcal{N}} (r_i-\bar{r})^2 \le \frac{1}{N}\sum_{i \in \mathcal{N}} r_i^2 \le \tfrac{L^2}{4}\rho^4$. Then, $\Delta_{\mathrm{FI}}(\theta,\rho)$ is upper bounded by
\begin{align} 
\mathbb{E}\big[\Delta_{\mathrm{FI}}(\theta,\rho)\big] \! \leq \!
\frac{3\rho^2}{N} \! \sum_{i \in \mathcal{N}} \mathbb{E} \! \left[ \|\delta_i(\theta)\|^2 \!+\! 4\|\zeta_{i,\phi_i}(\theta)\|^2 \right] \! +\! \frac{3}{4}L^2\rho^4,
\end{align}
which completes the proof.

\section{Proof of Lemma 2}\label{app:proof-lemma2}
Given batch $\phi_i$ sampled uniformly from $\mathcal{D}_i$, we have
\begin{align} \label{inner irrelevant}
\mathbb{E}\big[\langle \delta_i(\theta),\,\zeta_{i,\phi_i}(\theta)\rangle\big]
\!=\!
\mathbb{E}_i\Big[
\big\langle \delta_i(\theta),\mathbb{E}_{\phi_i\mid i}\big[\zeta_{i,\phi_i}(\theta)\big]\big\rangle \Big] \!
\overset{(a)}{=} \! 0,
\end{align}
where $(a)$ follows from $\mathbb{E}_{\phi_i\mid i}[\zeta_{i,\phi_i}(\theta)]=0$.

With $g_i-\nabla F(\theta)=\delta_i(\theta)+\zeta_{i,\phi_i}(\theta)$, we have
\begin{align}
\mathbb{E}\left[\big\|g_i \!-\! \nabla F(\theta)\big\|^2\right]
& \!=\!
\mathbb{E}\left[\big\|\delta_i(\theta)\big\|^2
\!+\! \big\|\zeta_{i,\phi_i}(\theta)\big\|^2 \right] 
\nonumber\\
& \!+\! \mathbb{E}\left[2\big\langle \delta_i(\theta),\,\zeta_{i,\phi_i}(\theta)\big\rangle\right], \nonumber \\
& \overset{(a)}{=} \mathbb{E}\Big[\!\big\|\delta_i(\theta)\big\|^2\Big]
\!+\! \mathbb{E}\Big[\!\big\|\zeta_{i,\phi_i}(\theta)\big\|^2\Big],
\end{align}
where $(a)$ comes from \eqref{inner irrelevant}. The proof is complete.

\section{Local Update Variance Suppression}\label{app:local-var}
With the variance-suppression operation in \eqref{vs local update}, the variance of $u_i^{t,k}$ is suppressed by $\gamma_l^2$ with $\gamma_l^2 < 1$.
Let $\mathcal{F}^{t,k}$ denote the $\sigma$-algebra~\cite{bain2009fundamentals} generated by $(\theta^t,h^t)$, $\mathcal{S}^t$, and all data sampled before training iteration $k$ of round $t$, i.e., $\{\phi_j^{t,\ell}\}_{j\in\mathcal{S}^t,\,0\le \ell\le k-1}$.
Given $\mathcal{F}^{t,k}$, $\theta_i^{t,k}$ and $h^t$ are fixed, and the remaining randomness comes from the current batch $\phi_i^{t,k}$, which determines $g_i^{t,k}$, $\tilde{\theta}_i^{t,k}$ through \eqref{vs flatness search}, and $\tilde{g}_i^{t,k}=\nabla F_i(\tilde{\theta}_i^{t,k};\phi_i^{t,k})$.
Conditioning on $\mathcal{F}^{t,k}$, we have
\begin{align} \label{vs_local_var_app}
&\mathbb{E}\!\left[u_i^{t,k}\mid\mathcal{F}^{t,k}\right]
=(1-\gamma_l)h^{t}+\gamma_l\,\mathbb{E}\!\left[\tilde g_i^{t,k}\mid\mathcal{F}^{t,k}\right]; \nonumber\\
&\mathrm{Var}\!\left[u_i^{t,k}\mid\mathcal{F}^{t,k}\right]
\overset{(a)}{=} \mathbb{E}\!\left[\left\|u_i^{t,k}-\mathbb{E}\!\left[u_i^{t,k}\mid\mathcal{F}^{t,k}\right]\right\|_2^2\middle|\mathcal{F}^{t,k}\right] \nonumber\\
&= \! \gamma_l^2\mathbb{E}\!\!\left[\!\left\|\tilde g_i^{t,k} \!\!-\! \mathbb{E}\!\!\left[\tilde g_i^{t,k} \!\! \mid \!\! \mathcal{F}^{t,k}\!\right]\!\right\|_2^2\middle|\mathcal{F}^{t,k} \! \right] \!\! = \! \gamma_l^2\mathrm{Var}\!\left[\!\tilde g_i^{t,k} \!\! \mid \!\! \mathcal{F}^{t,k}\!\right],
\end{align}
where $(a)$ follows from $\mathrm{Var}[X \! \mid \! \mathcal{F}] \triangleq \mathbb{E}[\|X-\mathbb{E}[X\mid\mathcal{F}]\|_2^2 \mid \mathcal{F}]$.

\section{Proof of Theorem 1}\label{app:proof-theorem1}

By the definition of the SAM flatness search, for each $\theta^t$,
let $\tilde{\theta}^t=\theta^t+\epsilon^t$ denote a perturbed model satisfying
$\|\epsilon^t\|\le \rho$ and $F(\tilde{\theta}^t)=\max_{\|\epsilon\|\le \rho} F(\theta^t+\epsilon)$.
With the smoothness of $F$, and by taking expectation of $F(\tilde{\theta}^{t+1})$
over the data and device-sampling randomness at round $t$, we have
\begin{align} \label{lipschitz of f}
    &\mathbb{E}[F(\theta^{t+1})] \leq \mathbb{E}[F(\tilde{\theta}^{t+1})] \nonumber \\
    &\leq F(\tilde{\theta}^t) \!+\! \mathbb{E} \left[\langle \nabla F(\tilde{\theta}^t), \tilde{\theta}^{t+1}-\tilde{\theta}^t \rangle \right] \!+\!  \frac{L}{2} \mathbb{E}\left[ \|\tilde{\theta}^{t+1}-\tilde{\theta}^t\|^2 \right] \nonumber \\
    & \overset{(a)}{\leq} \! F(\tilde{\theta}^t) \!-\! \eta_g \| \nabla F(\tilde{\theta}^t) \|^2 \!+\! \eta_g  \langle \nabla F(\tilde{\theta}^t), \nabla F(\tilde{\theta}^t) \!-\! h^{t+1} \rangle \!+\! a_1  \nonumber \\
    & \overset{(b)}{\leq} F(\tilde{\theta}^t) - \frac{\eta_g}{2} \| \nabla F(\tilde{\theta}^t) \|^2 + \frac{\eta_g}{2} \| \nabla F(\tilde{\theta}^t)- h^{t+1} \|^2  \nonumber \\
    & + L\eta_g^2 \Big(   \|   \nabla F(\tilde{\theta}^t) - h^{t+1} \|^2 + \| \nabla F(\tilde{\theta}^t)  \|^2 \Big)  \nonumber \\
    & \overset{(c)}{\leq} F(\tilde{\theta}^t) \!\!-\!\! \frac{11\eta_g}{24} \| \nabla F(\tilde{\theta}^t) \|^2 \!\!+\!\! \frac{13\eta_g}{24} \| \nabla F(\tilde{\theta}^t) \!-\! h^{t+1} \|^2 ,
\end{align}
where $a_1 = \frac{L\eta_g^2}{2} \mathbb{E}\left[ \| h^{t+1} \|^2 \right]$. Here, $(a)$ is due to \eqref{vs global update}, $(b)$ holds due to the Cauchy-Schwarz inequality, and $(c)$ comes from $\eta_g L \leq \frac{1}{24}$.

To deal with $\| \nabla F(\tilde{\theta}^t) - h^{t+1} \|^2$ in \eqref{lipschitz of f}, we first have
\begin{align} \label{A and B}
    & \mathbb{E} \! \left[ \| \nabla F (\tilde{\theta}^t )  - h^{t+1} \|^2 \right] \!=\! \mathbb{E}\left[ \| \nabla F (\tilde{\theta}^t ) \!-\! (1 \!-\! \gamma_l\gamma_g) h^{t} \!-\! a_2 \|^2 \right] \nonumber \\    
    &= \mathbb{E}\left[ \| \nabla F (\tilde{\theta}^t ) - (1-\gamma_l\gamma_g) h^{t} - a_3 +  (a_3-a_2) \|^2 \right] \nonumber \\
    & \overset{(a)}{=} \underbrace{ \mathbb{E}\left[ \| a_4 + \gamma_l \gamma_g\nabla F (\tilde{\theta}^t ) - a_3 \|^2 \right] }_A + \underbrace{ \mathbb{E}\left[ \| a_3-a_2 \|^2 \right] }_B , 
\end{align}
where $a_2 = \frac{\gamma_l \gamma_g}{S K} \sum_{i \in \mathcal{S}^t} \sum_{k=0}^{K-1} \tilde{g}_i^{t, k}$, $a_3 = \frac{\gamma_l \gamma_g}{N K} \sum_{i \in \mathcal{N} }\sum_{k=0}^{K-1} \tilde{g}_i^{t, k} $, and $a_4 = (1-\gamma_l\gamma_g)(\nabla F (\tilde{\theta}^t )  - h^{t})$. Moreover, $(a)$ is due to $\mathbb{E}[a_3-a_2]=0$.

Since $ \{ \tilde{g}_i^{t,k}  \}_{0 \leq k<K}$ are sequentially correlated, the upper bound of $A$ is derived as
\begin{align}
    & A = \mathbb{E}\left[ \| a_4 \|^2 \right] + (\gamma_l\gamma_g)^2\mathbb{E}\left[ \|  \nabla F (\tilde{\theta}^t ) - \frac{a_3}{\gamma_l\gamma_g}  \|^2 \right] \nonumber \\
    & + \mathbb{E}\left[ \left\langle  \sqrt{\gamma_l\gamma_g}a_4 , 2\sqrt{\gamma_l \gamma_g}(\nabla F (\tilde{\theta}^t ) -  \frac{a_5}{\gamma_l \gamma_g} \right\rangle \right] \nonumber \\
    & \overset{(a)}{\leq} (1+\frac{\gamma_l\gamma_g}{2}) \mathbb{E}\left[ \| a_4 \|^2 \right] + 2 \gamma_l\gamma_g \mathbb{E}\left[ \|  \nabla F (\tilde{\theta}^t ) -  \frac{a_5}{\gamma_l \gamma_g} \|^2 \right] \nonumber \\
    & + 2(\gamma_l\gamma_g)^2\mathbb{E}\Bigg[ \Big\|  \nabla F (\tilde{\theta}^t ) - \frac{a_5}{\gamma_l \gamma_g} \Big\|^2 + \frac{\sigma_{l,1}^2}{NK}\Bigg] \nonumber \\
    & \overset{(b)}{\leq} (1+\frac{\gamma_l\gamma_g}{2}) \mathbb{E}\left[ \| a_4 \|^2 \right] + 2 \gamma_l\gamma_g \mathbb{E}\left[ 2L^2 U^t + 8L^2 \rho^2 \right] \nonumber \\
    &  + 2(\gamma_l\gamma_g)^2\mathbb{E}\Big[ 2L^2 U^t + 8L^2 \rho^2 + \frac{\sigma_{l,1}^2}{NK}\Big] ,
\end{align}
where $U^t = \frac{1}{N K} \sum_{i \in \mathcal{N}} \sum_{k=0}^{K-1}\|\theta^t-\theta_i^{t, k}\|^2$, and $a_5 = \frac{\gamma_l \gamma_g}{N K} \sum_{i \in \mathcal{N}} \sum_{k=0}^{K-1} \nabla F_i (\tilde{\theta}_i^{t, k}) $. Here, $(a)$ is due to $2\langle \boldsymbol{x},\boldsymbol{y} \rangle \leq \|\boldsymbol{x}\|^2 + \|\boldsymbol{y}\|^2$, $ \mathbb{E}[a_3] = a_5$, and \textbf{Assumption 2}; and $(b)$ holds because of \textbf{Assumption 1}, the Cauchy-Schwarz inequality, and
\begin{align} \label{rho bound}
    & \|\tilde{\theta}^t - \tilde{\theta}_i^{t,k} \|^2 \leq 2\| {\theta}^t - {\theta}_i^{t,k} \|^2 + 2\| \epsilon^t - \epsilon_i^{t,k} \|^2 \nonumber \\
    & \leq 2\| {\theta}^t - {\theta}_i^{t,k} \|^2 + 2\rho^2 \cdot 4 = 2\| {\theta}^t - {\theta}_i^{t,k} \|^2 + 8\rho^2 . 
\end{align}

To bound $B$ in \eqref{A and B}, we have
\begin{align}
    & B \overset{(a)}{\leq} (\gamma_l\gamma_g)^2 \frac{N-S}{SN(N-1)} \sum_{i=1}^N \mathbb{E}\Big[ \Big\| \frac{1}{K} \sum_{k=0}^{K-1} \tilde{g}_i^{t, k} \!-\! \nabla F(\theta^t) \Big\|^2 \Big] \nonumber \\
    & \overset{(b)}{\leq} (\gamma_l\gamma_g)^2 \frac{N-S}{S(N-1)} \cdot \frac{3}{N} \sum_{i=1}^N \mathbb{E}\left[ \left\| a_6 \right\|^2 + \left\| a_7  \right\|^2  +  \left\| a_8  \right\|^2  \right] \nonumber \\
    & \overset{(c)}{\leq} (\gamma_l\gamma_g)^2 \frac{3(N \!-\! S)}{S(N \!-\! 1)} \mathbb{E}\left[ 2 L^2 U^t \!+\! 8 L^2 \rho^2 \!+\! \sigma_{l,1}^2 \!+\! \sigma_{g,1}^2 \right],
\end{align}
where $a_6 = \frac{1}{K} \sum_{k=0}^{K-1} \nabla F(\tilde{\theta}_i^{t, k})-\nabla F(\tilde{\theta}^t)$, $a_7 = \frac{1}{K} \sum_{k=0}^{K-1}(\nabla F_i(\tilde{\theta}_i^{t, k})-\nabla F(\tilde{\theta}_i^{t, k}))$, and $a_8 = \frac{1}{K} \sum_{k=0}^{K-1} (\tilde{g}_i^{t, k}-\nabla F_i(\tilde{\theta}_i^{t, k}))$;
Here, $(a)$ comes from $\frac{1}{N} \sum_{i \in \mathcal{N}}\| \boldsymbol{x}_i-\bar{{\boldsymbol{x}}}\|^2 \leq \frac{1}{N} \sum_{i \in \mathcal{N}}\|\boldsymbol{x}_i-\boldsymbol{y}\|^2$ with $\bar{{\boldsymbol{x}}} = \frac{1}{N}\sum_{i \in \mathcal{N}} \boldsymbol{x}_i$ and \cite[Lem. 7]{cheng2024momentum}; $(b)$ is due to the Cauchy-Schwarz inequality; and $(c)$ holds under \textbf{Assumptions 1-3}, and \eqref{rho bound}. 

With the upper bounds of $A$ and $B$, we follow the standard first-order treatment in SAM-based FL convergence analysis~\cite{qu2022generalized,pmlr-v235-fan24c} with
$\tilde{\theta}^t-\tilde{\theta}^{t-1}\approx {\theta}^t-{\theta}^{t-1}$ and obtain
\begin{align} \label{gradient difference temp 1}
    & \mathbb{E} \! \left[ \| \nabla F (\tilde{\theta}^t ) \!-\! h^{t+1} \|^2 \right] \overset{(a)}{\leq} (1 \!+\! \frac{\gamma_l\gamma_g}{2})(1 \!-\! \gamma_l\gamma_g)^2 \mathbb{E}\left[ (1 \!+\! \frac{\gamma_l\gamma_g}{2})\cdot \right.  \nonumber \\
    & \left. \|\nabla F(\tilde{\theta}^{t-1}) - h^t \|^2 + (1+\frac{2}{\gamma_l\gamma_g})L^2 \|{\theta}^t-{\theta}^{t-1}\|^2 \right] + C^t \nonumber\\
    & \overset{(b)}{\leq} (1+\frac{\gamma_l\gamma_g}{2})(1-\gamma_l\gamma_g)^2\mathbb{E}\left[ (1 \!+\! \frac{\gamma_l\gamma_g}{2})\|\nabla F(\tilde{\theta}^{t-1}) - h^t \|^2 + \right. \nonumber \\
    & \left. (1 \!+\! \frac{2}{\gamma_l\gamma_g})2L^2\eta_g^2 \left( \|h^t \!-\! \nabla F(\tilde{\theta}^{t-1})\|^2 \!+\! \|\nabla F(\tilde{\theta}^{t-1})\|^2 \right) \right] \!+\! C^t \nonumber\\
    & \overset{(c)}{\leq} \!\! \frac{9 \!-\! 8\gamma_l\gamma_g}{9} \!\! \|\nabla \! F(\tilde{\theta}^{t\!-\!1}) \!-\! h^t \|^2 \!\!+\!\! \frac{4L^2\eta_g^2}{\gamma_l\gamma_g} \! \|\!\nabla F \! (\tilde{\theta}^{t \!-\! 1})\!\|^2 \!+\! C^t,
\end{align}
where $C^t = 2(\gamma_l\gamma_g)^2 (2L^2 U^t + 8L^2 \rho^2 + \frac{\sigma_{l,1}^2}{NK} ) + 2\gamma_l\gamma_g(2L^2 U^t + 8L^2 \rho^2) + (\gamma_l\gamma_g)^2 \frac{3(N-S)}{S(N-1)} \mathbb{E}\left[ 2 L^2 U^t +8 L^2 \rho^2+ \sigma_{l,1}^2+\sigma_{g,1}^2 \right]$. Here, $(a)$ comes from \textbf{Assumption 1}, Young's inequality, and $\tilde{\theta}^t-\tilde{\theta}^{t-1} \approx {\theta}^t-{\theta}^{t-1}$; $(b)$ is due to the Cauchy-Schwarz inequality; and $(c)$ holds with $\eta_g L \leq \frac{\gamma_l\gamma_g}{6}$.

Similarly, by defining $\tilde{\theta}^{-1} = \tilde{\theta}^{0}$, we have
\begin{align}
    & \mathbb{E}\left[ \| \nabla F (\tilde{\theta}^0 ) \!- \! h^{1} \|^2 \right] \leq (1  \!- \! \gamma_l\gamma_g)\mathbb{E}  \! \left[ \| \nabla F (\tilde{\theta}^{-1}) \!- \! h^{0} \|^2 \right]  \!+ \! C^0 \nonumber \\
    & = (1-\gamma_l\gamma_g)\mathbb{E}\left[ \| \nabla F (\tilde{\theta}^{0} )  - h^{0} \|^2 \right]   + C^0.
\end{align}
Then, we derive the upper bound of $U^t$. 
Define $d_i^{t,k} = \mathbb{E}[ \theta_i^{t, k+1}-\theta_i^{t, k} \mid \mathcal{F}_i^{t, k} ] = -\eta_l ((1-\gamma_l) h^t+\gamma_l \nabla F_i(\tilde{\theta}_i^{t, k}))$.
For any $ 1 \leq j \leq k-1 \leq K-2 $, $\mathbb{E} [ \| d_i^{t,j} \|^2 ]$ is bounded as
\begin{align} \label{temp drift bound 1}
    & \mathbb{E} \! \Big[ \Big\| d_i^{t,j} \Big\|^2 \Big] \! \overset{(a)}{\leq} \!  \Big(1 \!\!+\!\! \frac{1}{k} \Big) \mathbb{E}\!\Big[ \Big\| d_i^{t,j\!-\!1} \Big\|^2 \Big] \!\!+\!\! (1 \!+\! k) \mathbb{E} \! \Big[\Big\| d_i^{t, j} \!-\! d_i^{t, j\!-\!1}\Big\|^2\Big] \nonumber \\
    & \overset{(b)}{\leq} \Big(1 \!+\! \frac{1}{k} \Big) \mathbb{E}\Big[ \Big\| d_i^{t,j-1} \Big\|^2 \Big] \nonumber \\
    & + (1+k)\eta_l^2\gamma_l^2 L^2 \mathbb{E}\left[2\left\| {\theta}_i^{t, j} - {\theta}_i^{t, j-1}  \right\|^2 + 8\rho^2 \right] \nonumber \\
    & \overset{(c)}{\leq} \Big(1+\frac{1}{k} +2(1+k)\eta_l^2\gamma_l^2 L^2 \Big)\mathbb{E}\Big[\Big\| d_i^{t, j-1}\Big\|^2\Big] \nonumber \\
    & + (1+k)\eta_l^2\gamma_l^2 L^2 ( 2\eta_l^2\gamma_l^2 \sigma_{l,1}^2 + 8\rho^2 ) \nonumber \\
    & \overset{(d)}{\leq} \Big(1 \!+\! \frac{2}{k} \Big)\mathbb{E}\Big[\Big\| d_i^{t, j-1}\Big\|^2\Big] \!+\! 2(1 \!+\! k)\eta_l^2\gamma_l^2 L^2 ( \eta_l^2\gamma_l^2 \sigma_{l,1}^2 \!+\! 4\rho^2 ) \nonumber \\
    & \overset{(e)}{\leq} e^2 \mathbb{E}\Big[\Big\| d_i^{t, 0}\Big\|^2\Big] + 6k^2\eta_l^2\gamma_l^2 L^2 ( \eta_l^2\gamma_l^2 \sigma_{l,1}^2 + 4\rho^2 ),
\end{align}
where $(a)$ is due to Young's inequality; $(b)$ comes from \textbf{Assumption 1}, the Cauchy-Schwarz inequality, and \eqref{rho bound}; $(c)$ holds under \textbf{Assumption 2}; $(d)$ comes from $\eta_l L \leq \frac{1}{\sqrt{2}\gamma_l K} \leq \frac{1}{\sqrt{2}\gamma_l(k+1)}$; and $(e)$ holds with $\left(1+\frac{2}{k}\right)^k \leq e^2$ and $(k+1)(1+\frac{2}{k})^k \leq 3k^2$ for $k \geq 2$.

With the bound in \eqref{temp drift bound 1}, we further obtain
\begin{align} \label{temp drift bound 2}
    & \mathbb{E}\Big[ \Big\| \theta_i^{t,k} - \theta^t \Big\|^2 \Big] 
    \overset{(a)}{\leq} k\sum_{j=0}^{k-1}\mathbb{E}\Big[\Big\| d_i^{t, j}\Big\|^2\Big]+ k \eta_l^2 \gamma_l^2 \sigma_{l,1}^2 \nonumber \\
    & \overset{(b)}{\leq}  \frac{k^2}{2}(e^2-1) \mathbb{E}\Big[\Big\|d_i^{t, 0}\Big\|^2\Big]+\frac{k^3}{2}(e^2-3)(1+k) \eta_l^2 \gamma_l^2 L^2 \cdot \nonumber \\
    & \Big(\eta_l^2 \gamma_l^2 \sigma_{l, 1}^2+4 \rho^2\Big)  +k \eta_l^2 \gamma_l^2 \sigma_{l, 1}^2 ,
\end{align}
where $(a)$ is due to the Cauchy-Schwarz inequality and \textbf{Assumption 2}; and $(b)$ comes from recursion from \eqref{temp drift bound 1}.

Summing \eqref{temp drift bound 2} over $i$ and $k$,  we bound $U^t$ as
\begin{align} \label{bound of U^t}
    & U^t \!=\! \frac{1}{NK}\sum_{i \in \mathcal{N}} \sum_{k=0}^{K-1}\mathbb{E}\left[ \left\| \theta_i^{t,k} \!-\! \theta^t \right\|^2 \right] \overset{(a)}{\leq} \frac{e K^2}{2N} \! \sum_{i \in \mathcal{N}} \mathbb{E} \! \left[\|d_i^{t, 0}\|^2\right] \nonumber \\
    &  + \frac{K}{2}\eta_l^2 \gamma_l^2 \sigma_{l, 1}^2 + \frac{e K^4}{2}\eta_l^2 \gamma_l^2 L^2\left(\eta_l^2 \gamma_l^2 \sigma_{l, 1}^2+4 \rho^2\right) ,
\end{align}
where $(a)$ is obtained by substituting \eqref{temp drift bound 2} into $U^t$, summing over $k$ and $i$, and bounding the resulting polynomial terms concerning $k$.

Next, we derive the bound of $d^t \!= \frac{1}{N} \! \sum_{i \in \mathcal{N}} \mathbb{E}[\|d_i^{t, 0}\|^2]$. Recalling that $d_i^{t,0} \!=\! -\eta_l \left(\left(1-\gamma_l\right) h^t+\gamma_l \nabla F_i(\tilde{\theta}_i^{t})\right)$, we have
\begin{align}
    \frac{1}{N} \!\! \sum_{i \in \mathcal{N}} \! \mathbb{E} \! \Big[ \! \|d_i^{t, 0}\|^2 \! \Big] \!\!\leq\!\! 2\eta_l^2 \! \Big( \! \|(1 \!\!-\!\! \gamma_l)h^t\!\|^2 \!\!+\!\!  \frac{\gamma_l^2}{N} \!\! \sum_{i \in \mathcal{N}} \! \| \nabla \! F_i(\tilde{\theta}_i^t) \! \|^2 \! \Big),
\end{align}
For the upper bound of $\| \nabla F_i(\tilde{\theta}_i^t) \|^2$, we have
\begin{align} \label{bound of f_i theta_i}
    &\mathbb{E}[\|\nabla F_i(\tilde{\theta}^t)\|^2] 
    \overset{(a)}{\leq} (1\!+\!q) \mathbb{E}[\|\nabla F_i(\tilde{\theta}^{t-1})\|^2] + 2(1\!+\!\frac{1}{q}) \eta_g^2 L^2 \cdot \nonumber \\
    & ( \mathbb{E}[\|h^t \!\!-\!\! \nabla \! F \! (\tilde{\theta}^{t\!-\!1})\|^2] \!\!+\!\!\mathbb{E}[\|\nabla \! F \! (\tilde{\theta}^{t\!-\!1})\|^2]) \!\!\overset{(b)}{\leq}\!\! (1\!\!+\!\!q)^t \mathbb{E}[\|\nabla \!F_i  (\tilde{\theta}^0)\|^2] \nonumber \\
    & + \!\! \frac{2}{q} \eta_g^2 L^2 \! \sum_{j=0}^{t-1}(\mathbb{E}[\|h^{j\!+\!1} \!\!-\!\! \nabla F(\tilde{\theta}^{j})\|^2]\!+\!\mathbb{E}[\|\nabla \! F\!(\tilde{\theta}^j)\|]^2)(1\!+\!q)^{t-j} \nonumber \\
    & \overset{(c)}{\leq} \! e \mathbb{E}[\|\nabla F_i(\tilde{\theta}^0) \|^2] \!+\! 2 e(t \!+\! 1) \eta_g^2 L^2 \! \sum_{j=0}^{t-1} Q^j,
\end{align}
where $Q^j = \mathbb{E}[\|h^{j + 1} \!-\! \nabla F(\tilde{\theta}^{j})\|^2 \!+\! \|\nabla F(\tilde{\theta}^j)\|^2 ]$; $(a)$ comes from Young's inequality, \textbf{Assumption 1}, and the Cauchy-Schwarz inequality; 
$(b)$ follows by recursively applying the recursion in $(a)$ from $0$ to $t-1$;
and $(c)$ is obtained by setting $q=\frac{1}{t}$.

The upper bound of $\sum_{t=0}^{T-1} d^t = \sum_{t=0}^{T-1} \frac{1}{N} \! \sum_{i \in \mathcal{N}} \mathbb{E}[\|d_i^{t, 0}\|^2]$ is then derived as 
\begin{align} \label{bound of sum d^t}
    & \sum_{t=0}^{T-1} \! d^t 
    \! \overset{(a)}{\leq} \! 4\eta_l^2(1 \!-\! \gamma_l)^2 \! \sum_{t=0}^{T-1} \! \Big( \| \nabla F(\tilde{\theta}^{t \!-\! 1}) \!-\! h^t \|^2 \!+\! \| \nabla F(\tilde{\theta}^{t \!-\! 1}) \|^2 \Big) \nonumber \\
    & + 2\eta_l^2\gamma_l^2 e T \frac{1}{N}\sum_{i \in \mathcal{N}} \|\nabla F_i(\tilde{\theta}^0)\|^2 + 2 e \eta_l^2\eta_g^2 \gamma_l^2 L^2 T(T+1) \cdot \nonumber \\
    &\sum_{j=0}^{T-2} (\mathbb{E}[\|\nabla F(\tilde{\theta}^{j}) - h^{j+1} \|^2]+\mathbb{E}[\|\nabla F(\tilde{\theta}^j)\|^2]) \nonumber \\
    & \overset{(b)}{\leq} \! \frac{1}{\delta e K^2 L^2} \!\! \sum_{t=-1}^{T-2} \! \Big( \! \| \nabla F(\tilde{\theta}^{t}) \!-\! h^{t\!+\!1} \|^2 \!+\! \| \nabla F(\tilde{\theta}^{t}) \|^2 \! \Big) \nonumber \\
    & + 2\eta_l^2\gamma_l^2 e T \frac{1}{N}\sum_{i \in \mathcal{N}} \|\nabla F_i(\tilde{\theta}^0)\|^2 ,
\end{align}
where $(a)$ comes from the Cauchy-Schwarz inequality and \eqref{bound of f_i theta_i}; and $(b)$ holds by letting $\delta e K^2 L^2 (4 \eta_l^2\left(1-\gamma_l\right)^2 + 2 e \eta_l^2 \eta_g^2 \gamma_l^2 L^2 T(T+1) ) \leq 1$ with $\delta > 0$.

With the bound of $U^t$ derived in \eqref{bound of U^t}, \eqref{gradient difference temp 1} is bounded as
\begin{align}
    & \mathbb{E}\Big[ \| \nabla F (\tilde{\theta}^t )  \!- \! h^{t+1} \|^2 \Big] 
    \!\leq\! (1 \!-\! \frac{8\gamma_l\gamma_g}{9})\|\nabla F(\tilde{\theta}^{t-1}) \!-\! h^t \|^2  \\
    & + \! \frac{4L^2\eta_g^2}{\gamma_l\gamma_g}\mathbb{E}\Big[ \|\nabla F(\tilde{\theta}^{t \!-\! 1})\|^2 \Big] \!+\! a_9 \frac{e K^2}{2N} \sum_{i \in \mathcal{N}} \mathbb{E}\Big[\Big\|d_i^{t, 0}\Big\|^2\Big] \!+\! a_{10} \nonumber \\
    & + \! a_9 \! \Big( \frac{K}{2} \eta_l^2 \gamma_l^2 \sigma_{l, 1}^2 \!+\! \frac{e K^4}{2} \eta_l^2 \gamma_l^2 L^2 (\eta_l^2 \gamma_l^2 \sigma_{l, 1}^2 \!+\! 4 \rho^2) \Big) \!+\! a_{11}  , \nonumber
\end{align}
where $a_9 = 2\gamma_l\gamma_g L^2 (2\gamma_l\gamma_g + 2 + 3\gamma_l\gamma_g\frac{N-S}{S(N-1)}) $, $a_{10} = 8L^2\rho^2\gamma_l\gamma_g(2\gamma_l\gamma_g+2+3\gamma_l\gamma_g\frac{N-S}{S(N-1)})$, and $a_{11} = (\gamma_l\gamma_g)^2(\frac{2\sigma_{l,1}^2}{NK} + 3\frac{N-S}{S(N-1)}(\sigma_{l,1}^2+\sigma_{g,1}^2) )$. Meanwhile, we have
\begin{align}
    & \mathbb{E} \! \Big[ \! \| \nabla \! F \! (\tilde{\theta}^0 ) \!-\! h^{1} \|^2 \! \Big] \!\!\leq \!\! (1 \!-\! \gamma_l\gamma_g)\mathbb{E}\Big[ \! \| \nabla F \!(\tilde{\theta}^{-1} ) \!-\! h^{0} \|^2 \! \Big]  \!\! + \! C^0 \\
    & \overset{(a)}{\leq} \!\! (1 \!\!-\!\gamma_l\gamma_g)\mathbb{E}\!\Big[ \! \| \nabla \! F \! (\tilde{\theta}^{-1} ) \!\!-\!\! h^{0} \|^2 \! \Big] \!\!+\! a_9 \frac{e K^2}{2N} \!\! \sum_{i \in \mathcal{N}} \!\! \mathbb{E} \! \Big[ \Big\|d_i^{0, 0}\Big\|^2\Big] \!\!+\! a_{12} ,  \nonumber
\end{align}
where $a_{12} = a_9 \Big(\frac{K}{2} \eta_l^2 \gamma_l^2 \sigma_{l, 1}^2+e \frac{K^4}{2} \eta_l^2 \gamma_l^2 L^2 (\eta_l^2 \gamma_l^2 \sigma_{l, 1}^2+4 \rho^2 )\Big)+a_{10}+a_{11} $. Here, $(a)$ comes from the bound of $U^t$.

Define $\mathcal{E}^t = \mathbb{E}[ \| \nabla F (\tilde{\theta}^t )  - h^{t+1} \|^2 ]$.
Summing $\mathcal{E}^t$ over $t$ from 0 to $T\!\!-\!1$ and applying \eqref{bound of sum d^t}, we obtain
\begin{align} \label{convergence temp 1}
    &\sum_{t=0}^{T-1} \! \mathcal{E}^t \!\!
     =\!\! \frac{9 \!-\! 8\gamma_l \gamma_g}{9} \!\!\! \sum_{t=-1}^{T-2} \!\! \mathcal{E}^t \!\!+\!\! \frac{4 L^2 \eta_g^2}{\gamma_l \gamma_g} \!\! \sum_{t=0}^{T-2} \! \| \nabla \!F \! (\tilde{\theta}^{t}) \! \|^2 \!\!+\!\! \frac{a_9 e K^2}{2} \!\! \sum_{t=0}^{T-1} d^t \nonumber \\
     & \!\!+\!\! T a_{12} \!\! \overset{(a)}{\leq} \!\! \frac{9 \!-\! 8\gamma_l \gamma_g}{9} \!\!\! \sum_{t=-1}^{T-2} \!\! \mathcal{E}^t \!\!+\!\! \frac{4 L^2 \eta_g^2}{\gamma_l \gamma_g} \!\! \sum_{t=0}^{T-2} \! \|  \nabla \!F\! (\tilde{\theta}^{t}) \! \|^2 \!\!+\!\! 
    \frac{7\gamma_l\gamma_g}{\delta} \!\!\! \sum_{t=-1}^{T-2} \!\! (    \mathcal{E}^t  \nonumber \\
    & \!\!+\!\! \|\nabla \! F \! (\tilde{\theta}^t ) \|^2 ) \!\!+\!\! 14 \gamma_l^3 \gamma_g T (e\eta_l KL)^2 \! \frac{1}{N} \! \sum_{i \in \mathcal{N}} \!\! \|\nabla F_i(\tilde{\theta}^0)\|^2 \!+\! T a_{12} \nonumber \\
    & \overset{(b)}{\leq} \frac{9 - 7\gamma_l\gamma_g}{9}\sum_{t=-1}^{T-2} \mathcal{E}^t + \frac{2\gamma_l\gamma_g}{9}  \sum_{t=0}^{T-2} \|\nabla F(\tilde{\theta}^{t}) \|^2 \nonumber \\
    & + 14 \gamma_l^3 \gamma_g T (e\eta_l KL)^2 \frac{1}{N} \sum_{i \in \mathcal{N}} \|\nabla F_i (\tilde{\theta}^0 ) \|^2 + T a_{12} ,
\end{align}
where $(a)$ is due to $a_9 \leq 14\gamma_l\gamma_g L^2$, and $(b)$ is obtained by setting $\delta = 63$ and $\eta_g \leq \frac{\gamma_l \gamma_g}{6L}$.
By combining the like terms in \eqref{convergence temp 1}, we have
\begin{align} \label{convergence temp 2}
    & \sum_{t=0}^{T-1} \mathcal{E}^t \leq \frac{9}{7 \gamma_l\gamma_g} \mathcal{E}_{-1} \!+\! \frac{2}{7} \mathbb{E}\Big[\sum_{t=-1}^{T-2} \|\nabla F (\tilde{\theta}^t ) \|^2\Big] \!+\! \frac{9}{7\gamma_l\gamma_g}T a_{12} \nonumber \\
    & + \frac{126}{7}T \big(e \gamma_l \eta_l K L \big)^2 \frac{1}{N} \sum_{i \in \mathcal{N}} \|\nabla F_i (\tilde{\theta}^0 ) \|^2 ,
\end{align}

With $\frac{1}{\eta_g}\mathbb{E}[F(\tilde{\theta}^{t+1}) \!-\! F(\tilde{\theta}^t) ]
    \leq  \!-\! \frac{11}{24} \| \nabla F(\tilde{\theta}^t) \|^2 \!+\! \frac{13}{24} \|- h^{t+1} + \nabla F(\tilde{\theta}^t) \|^2$ revealed in \eqref{lipschitz of f}, by summing over $t$, we have
\begin{align} \label{convergence temp 3}
    & \frac{1}{\eta_g} \mathbb{E} \Big[ F(\tilde{\theta}^T) \!-\! F(\tilde{\theta}^0) \Big] \!\leq\! -\frac{8}{25} \! \sum_{t=0}^{T-1} \mathbb{E}\big[\|\nabla F (\tilde{\theta}^t ) \|^2 \big] \\
    & + \! \frac{39}{56\gamma_l\gamma_g} (\mathcal{E}^{-1} \!+\! a_{12}T) \!\!+\!\! \big(e \gamma_l \eta_l K L \big)^2 \frac{39T}{4N} \! \sum_{i \in \mathcal{N}} \! \|\nabla F_i (\tilde{\theta}^0) \|^2 , \nonumber
\end{align}

Under \textbf{Assumption 1} and $h^0=0$, we have $\mathcal{E}^{-1}=\mathbb{E}\!\left[\|\nabla F(\tilde{\theta}^0)-h^0\|^2\right]\leq 2L(F(\tilde{\theta}^0)-F^*) := 2L\Delta$. Defining $G^0 = \frac{1}{N} \sum_{i \in \mathcal{N}}\|\nabla F_i(\tilde{\theta}^0)\|^2$ and reorganizing \eqref{convergence temp 3}, we have
\begin{align} \label{convergence of fedvssam temp}
    \frac{1}{T} \!\! \sum_{t=0}^{T-1}  \mathbb{E}\!\Big[\big\|\nabla \! F\big(\tilde{\theta}^t\big)\big\|^2\Big] & \! \leq \! 
    \frac{(125\gamma_l \gamma_g \!+\! 88\eta_g L)\Delta}{20\eta_g \gamma_l \gamma_g T} \!+\! \frac{11}{5\gamma_l\gamma_g } a_{12} \nonumber \\
    & + \frac{31 e^2}{5} ( \gamma_l \eta_l K L)^2 G^0,
\end{align}
where
\begin{align} \label{app:eq:a12}
a_{12}
&= \gamma_l \gamma_g L^2\!\left(2+2 \gamma_l \gamma_g+3 \gamma_l \gamma_g \frac{N-S}{S(N-1)}\right)
\sigma_{l, 1}^2 \eta_l^2 \gamma_l^2\!\left(K+e K^4 \eta_l^2 \gamma_l^2 L^2\right) \nonumber\\
&\quad +4 \gamma_l \gamma_g L^2\!\left(2+2 \gamma_l \gamma_g+3 \gamma_l \gamma_g \frac{N-S}{S(N-1)}\right)
\rho^2\!\left(2+e K^4 \eta_l^2 \gamma_l^2 L^2\right) \nonumber\\
&\quad + (\gamma_l \gamma_g)^2\!\left[\frac{2 \sigma_{l, 1}^2}{N K}+3 \frac{N-S}{S(N-1)}(\sigma_{l, 1}^2+\sigma_{g, 1}^2)\right].
\end{align}
This completes the proof.

\section{Proof and Detailed Discussion of Theorem 2}\label{app:proof-theorem2}

\subsection{Proof of Theorem 2} \label{proof-theorem2 subsec}
Let $Q^t:=\mathbb{E} [\|h^{t}-\nabla F(\theta^{t})\|^2 ]$ and $E^t:=\mathbb{E} [\|h^{t}-\nabla F(\tilde{\theta}^{t-1})\|^2 ]$ for $t\ge 1$. We have
\begin{align}
    & Q^t \overset{(a)}{\leq} \! 2\mathbb{E}\!\left[\|h^t \!-\! \nabla F(\theta^{t-1})\|^2\right] \!+\! 2\mathbb{E}\!\left[\|\nabla F(\theta^{t-1}) \!-\! \nabla F(\theta^{t})\|^2\right] \label{eq:Qt_step0} \nonumber \\
    & \overset{(b)}{\leq} \! 4 E^t \!+\! 4\mathbb{E}\!\Big[\|\nabla F(\tilde{\theta}^{t-1}) \!-\! \nabla F(\theta^{t-1})\|^2 \Big] \!+\! 2L^2 \eta_g^2 \mathbb{E}\big[\|h^t\|^2 \big] \nonumber \\
    & \overset{(c)}{\leq} 4 E^t \!+\! 4L^2 \rho^2
    \!+\! 4L^2 \eta_g^2 (\|h^t \!-\! \nabla F(\theta^{t-1})\|^2 \!+\! \|\nabla F(\theta^{t-1})\|^2)
    \nonumber \\
    & \overset{(d)}{\leq} 4 E^t \!\!+\! 4L^2 \rho^2
    \!+\! 8L^2 \eta_g^2 (E^t \!+\! 2L^2\rho^2 \!+\! \|\nabla F(\tilde{\theta}^{t-1}) \|^2) \!=\! (4 + \nonumber \\
    & \!+\! 8L^2\eta_g^2 ) \! E^t 
  \!\!+\! 8L^2\eta_g^2 \|\nabla F(\tilde{\theta}^{t-1})\|^2 \!+\! \big(4L^2 \!\!+\!\! 16L^4\eta_g^2\big)\rho^2,
\end{align}
where $(a)$ is due to the Cauchy-Schwarz inequality, and $(b)$-$(d)$ come from the Cauchy-Schwarz inequality, \textbf{Assumption 1}, and $\|\epsilon^{t-1}\| \leq \rho$.

Summing $Q^t$ over $t$ from $0$ to $T-1$, we obtain
\begin{align}\label{eq:Qt_avg_1}
\frac{1}{T}\!\!\sum_{t=1}^{T} \! Q^t
\! \leq \! \frac{a_{13}}{T}\!\sum_{t=1}^{T} \!\! E_t \!+\! \frac{8L^2\eta_g^2}{T} \! \sum_{t=1}^{T}\!\mathbb{E}\!\big[\|\nabla \! F\!(\tilde{\theta}^{t \!-\! 1})\|^2\big] \!\!+\! a_{14},
\end{align}
where $a_{13} = 4 + 8L^2\eta_g^2$, and $a_{14} = (4L^2 + 16L^4\eta_g^2)\rho^2$.

Based on the definitions and parameter settings in \textbf{Corollary 1}, it readily follows that
\begin{align}
\label{eq:grad_avg}
\frac{1}{T}\sum_{t=0}^{T-1}\mathbb{E}\!\left[\|\nabla F(\tilde{\theta}^{t})\|^2\right]
\;\lesssim\;
\frac{L\Delta}{T} \;+\; \sqrt{\frac{L\Delta\,\Sigma}{T}} .
\end{align}
Here, $\lesssim$ hides a universal constant that does not depend on $T$, $K$, and all round-dependent terms. Based on \eqref{convergence temp 2}, the recursion of $E^t, \forall t$, is given as 
\begin{align} \label{eq:Et_avg}
\frac{1}{T}\!\!\sum_{t=1}^{T} \! E^t
\!\! \lesssim \!\!
\frac{L\Delta}{\beta T}
\!\!+\!\!
\frac{1}{T}\!\!\sum_{t=0}^{T-1}\!\! \|\nabla F(\tilde{\theta}^{t})\|^2
\!\!+\!\!
(\gamma_l\eta_l K L)^2 G^0
\!\!+\!
\frac{a_{12}}{\beta},
\end{align}
where $(\gamma_l\eta_l K L)^2 G^0 \lesssim \frac{1}{\sqrt{T}}$, $\rho^2 \lesssim \frac{1}{L^2\sqrt{S K T}} \lesssim \frac{\Sigma}{\sqrt{T}}$, and $\frac{a_{12}}{\beta} \lesssim \beta\,\Sigma + \frac{\Sigma}{\sqrt{T}}$.

Substituting \eqref{eq:grad_avg} and \eqref{eq:Et_avg} into \eqref{eq:Qt_avg_1} yields
\begin{align}
\frac{1}{T}\sum_{t=1}^{T} Q^t
\lesssim
\frac{L\Delta}{\beta\,T}
+
\beta \Sigma
+
\sqrt{\frac{L\Delta\,\Sigma}{T}}
+
\frac{\Sigma}{\sqrt{T}},
\end{align}
which completes this proof.

\subsection{Detailed Discussion of Theorem 2} \label{discussion-theorem2 subsec}
Under the settings of \textbf{Corollary~\ref{convergence corollary}}, balancing $\frac{L\Delta}{\beta T} $ and $ \beta \Sigma$ with $\beta = \min\{1,\sqrt{L\Delta/(\Sigma T)}\}$ yields
\begin{align} \label{the good of variance suppression eq modified}
    \frac{1}{T}\sum_{t=1}^{T}\mathbb{E}\!\left[\|h^t-\nabla F(\theta^t)\|^2\right]\lesssim\sqrt{\frac{L\Delta\,\Sigma}{T}}+\frac{\Sigma}{\sqrt{T}}.
\end{align}
In other words, increasing the device participation, $S/N$, and controlling data heterogeneity can effectively tighten the MSE.
Recall the definition of $\Sigma = \frac{\sigma_{l,1}^2}{N K} + \frac{N-S}{S(N-1)}(\sigma_{l,1}^2+\sigma_{g,1}^2)$ in \textbf{Corollary 1}. When $\Sigma$ becomes large with a small $S$ or large $\sigma_{l, 1}^2$ and $\sigma_{g, 1}^2$, the MSE bound \eqref{the good of variance suppression eq} is increasingly dominated by $\beta \Sigma$, leading to a smaller $\beta$ setting, which reduces residual variance $\beta \Sigma$ while decelerating convergence, as discussed above.
In contrast, under a deterministic full device participation and IID data settings, the residual variance takes $\Sigma \to 0$, and the RHS of \eqref{the good of variance suppression eq} reduces to $\frac{L \Delta}{\beta T}$. 
In this case, setting $\beta = 1$ is optimal, under which the bound in \eqref{the good of variance suppression eq} reduces to the
MSE decay rate of $\mathcal{O}(L\Delta/T)$.

\end{document}